\documentclass[letterpaper]{article} 
\usepackage{aaai2027} 
\usepackage{times}  
\usepackage{helvet}  
\usepackage{courier}  
\usepackage[hyphens]{url}  
\usepackage{graphicx} 
\usepackage{natbib}  
\usepackage{caption} 
\usepackage{soul}

\usepackage{array}
\usepackage{makecell}
\usepackage{multirow}
\usepackage{booktabs} 
\usepackage{amsmath}
\usepackage{amssymb}  
\usepackage{xcolor}   
\usepackage{pifont}
\usepackage{colortbl} 
\usepackage{makecell}
\usepackage{tabularx}
\usepackage[most]{tcolorbox}
\definecolor{lightbluebg}{RGB}{225,240,255}

\newtcbox{\bluebadge}{
  on line,
  boxrule=0pt,
  colback=lightbluebg,
  colframe=lightbluebg,
  arc=2pt,
  boxsep=1pt,
  left=2pt,
  right=2pt,
  top=1pt,
  bottom=1pt
}
\definecolor{lightgreenbg}{RGB}{225,245,230}

\newtcbox{\greenbudge}{
  on line,
  boxrule=0pt,
  colback=lightgreenbg,
  colframe=lightgreenbg,
  arc=2pt,
  boxsep=1pt,
  left=2pt,
  right=2pt,
  top=1pt,
  bottom=1pt
}
\newcommand{\cmark}{\textcolor{green!60!black}{\ding{51}}} 
\newcommand{\xmark}{\textcolor{red}{\ding{55}}}            

\title{\raisebox{-0.7ex}{\includegraphics[height=5ex]{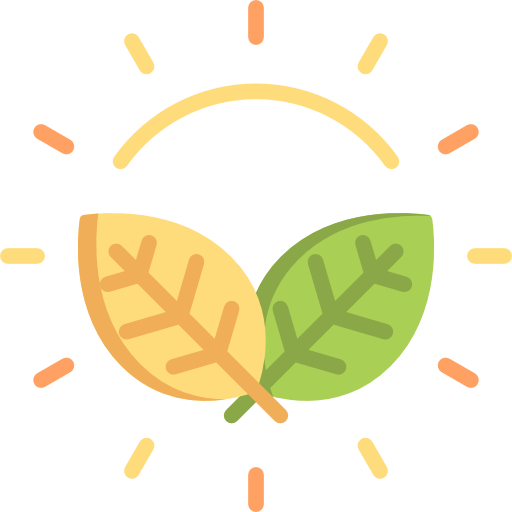}}HealBench: A Fine-Grained Benchmark for Evaluating Psychological Comfort in Text-to-Image Generation}

\author{
    Junjie Xu\textsuperscript{1},
    Yuzhe Yang\textsuperscript{1},
    Rui Yang\textsuperscript{2},
    Luwei Xiao\textsuperscript{3},\\
    Wenxin Hu\textsuperscript{1},
    Xingjiao Wu\textsuperscript{1},
    Liang He\textsuperscript{1}
}

\affiliations{
    \textsuperscript{1}East China Normal University\\
    \textsuperscript{2}Johns Hopkins University\\
    \textsuperscript{3}National University of Singapore
}

\begin{document}

\maketitle

\begin{abstract}
Current text-to-image and image-editing models can generate photorealistic images with strong semantic fidelity and high visual quality. Yet an important human-centered capability remains underexplored: whether these systems can create images that users perceive as psychologically comforting. Drawing on cognitive evaluation theory, we conceptualize psychological comfort not as a low-level visual property, but as an affective response shaped by a viewer’s appraisal of restorative and emotionally supportive cues.
To study this problem, we introduce HealBench, a large-scale, fine-grained benchmark for evaluating perceived psychological comfort in restorative visual stimuli. HealBench contains 30K high-quality image-text pairs spanning 19 high-level categories and 92 subcategories, constructed from 100K candidates through a human-and-expert-in-the-loop pipeline. To preserve the restorative nature of the data, the pipeline integrates VLM-based semantic matching, multiple human preference models, expert filtering, and augmentation guided by highly restorative exemplars. We further construct HealBench-Eval, a rigorously selected evaluation set of 1,095 instances using feature clustering.
Using HealBench-Eval, we conduct extensive objective and subjective evaluations of 14 representative image generation models, together with prompt-rewriting generation strategies. Our results reveal a clear gap between conventional visual generation quality and perceived psychological comfort: models that perform well on semantic alignment and aesthetics still struggle to produce images experienced as emotionally restorative. Further reinforcement learning experiments show that HealScorer can serve as a practical reward signal for assessing and improving the emotional and restorative quality of generated images. These findings suggest that psychological comfort is a distinct and underexplored dimension of human-centered image generation, and establish HealBench as a foundation for future work on affect-aware generative models.
\end{abstract}

\noindent\textbf{Keywords:} Text-to-Image, Emotional Comfort, Benchmark and datasets, Image Manipulation, Multi-modal Learning

\definecolor{bad}{RGB}{255, 46, 88}
\definecolor{good}{RGB}{18, 159, 88}
\definecolor{add}{RGB}{252,228,214}
\definecolor{name}{RGB}{0, 34, 255} 

\definecolor{DeepBlue}{RGB}{35, 70, 120}
\section{Introduction}
\label{sec:intro}
Recently, text-to-image (T2I) generation and instruction-guided image editing have made rapid progress in semantic alignment \cite{fuchats}, visual quality, editing fidelity \cite{yu2025anyedit, fu2025feededit}, and human preference modeling \cite{ma2025hpsv3}. These advances have enabled generative models to produce images that are increasingly faithful, realistic, and visually appealing. However, as visual generation systems are increasingly used in emotionally sensitive and human-centered scenarios, a more subtle question emerges: can a generated image not only \textbf{look correct}, but also \textbf{feel right} to the viewer? This distinction matters when images are expected to support emotional experience, restoration, or psychological relief rather than merely satisfy semantic or aesthetic criteria. An image may match the prompt and appear visually polished, yet still fail to provide \underline{perceived psychological comfort} \cite{belanche2025customer}.

\begin{figure}
    \centering
    \includegraphics[width=0.98\linewidth]{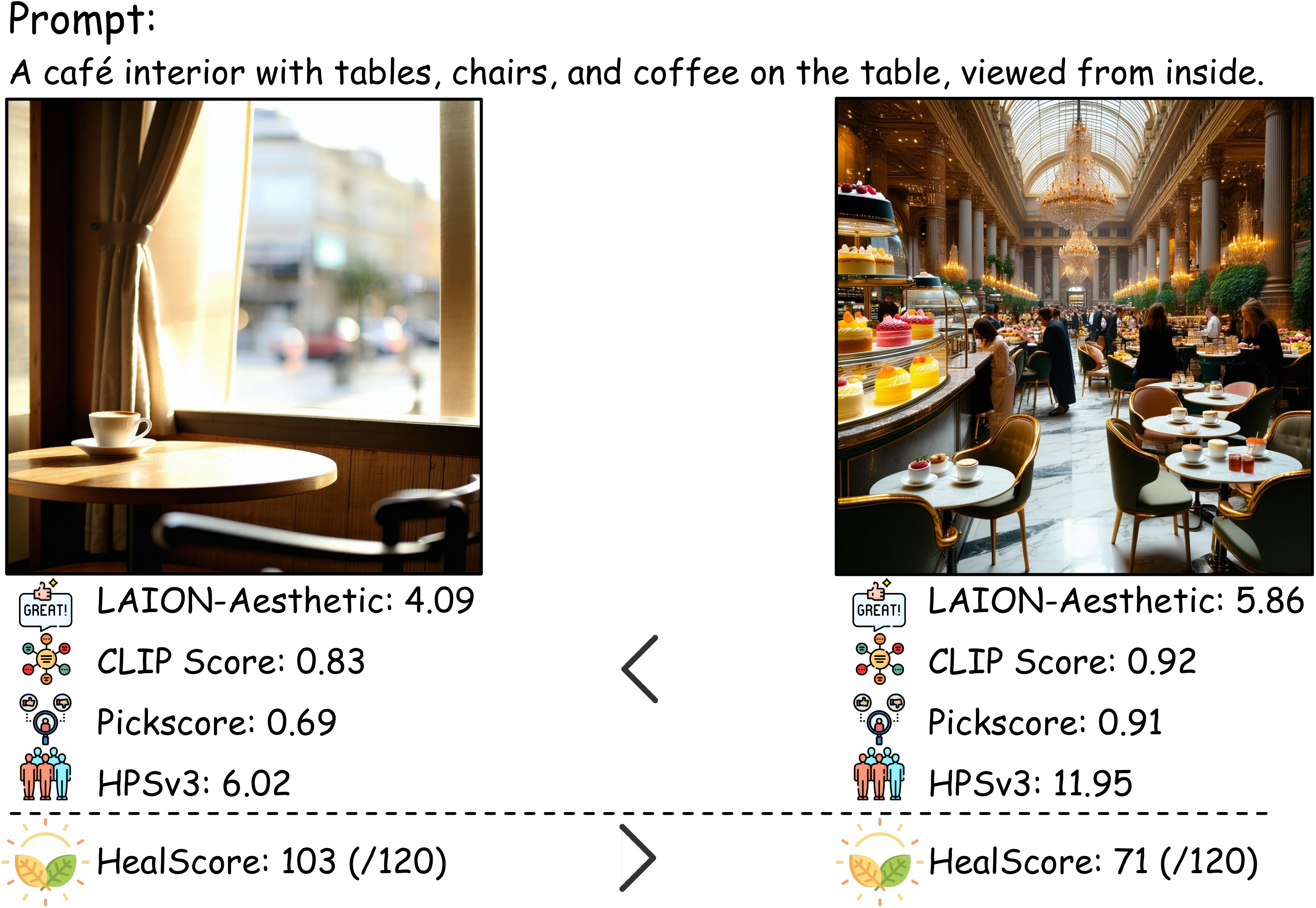}
    \vspace{-10pt}
    \caption{Despite sharing a prompt class, the left image scores poorly on general models (like CLIP) but high in psychological comfort (HealScore). Conversely, the right image has high evaluation scores but low comfort.}
    \label{fig:intro2}
    \vspace{-20pt}
\end{figure}

\begin{table*}[t]
\centering
\tiny
\setlength{\tabcolsep}{2.2pt}
\renewcommand{\arraystretch}{1.03}
\caption{Comparison with representative benchmarks and evaluators for T2I generation and image editing. Existing resources mainly focus on preference, alignment, fidelity, compositionality, or generic quality assessment, whereas \textbf{HealBench} explicitly targets \textbf{perceived psychological comfort} with \textbf{fine-grained diagnostic dimensions} and \textbf{actionable rationale-bearing feedback}.}
\vspace{-10pt}
\label{tab:benchmark_comparison}

\resizebox{\textwidth}{!}{%
\begin{tabular}{p{2.7cm} >{\centering\arraybackslash}p{1.05cm} >{\centering\arraybackslash}p{1.45cm} >{\centering\arraybackslash}p{4.7cm} >{\centering\arraybackslash}p{0.85cm} >{\centering\arraybackslash}p{1.0cm} >{\centering\arraybackslash}p{1.35cm} >{\centering\arraybackslash}p{1.0cm}}
\toprule
\textbf{Benchmark / Evaluator}
& \textbf{Venue}
& \textbf{Scope}
& \textbf{Primary Target}
& \makecell{\textbf{Fine-}\\\textbf{grained}}
& \makecell{\textbf{Comfort-}\\\textbf{specific}}
& \makecell{\textbf{Rationale /}\\\textbf{Feedback}}
& \makecell{\textbf{Optimi-}\\\textbf{zation}} \\
\midrule
\makecell[l]{Pick-a-Pic\\}
& NeurIPS'23
& T2I
& Pairwise human preference
& No \xmark
& No \xmark
& No \xmark
& Limited \\

\makecell[l]{AIGCIQA2023\\}
& LNCS'24
& T2I
& Quality, authenticity, and correspondence
& Partial
& No \xmark
& No \xmark
& No \xmark \\

\makecell[l]{T2I-CompBench\\}
& NeurIPS'23
& T2I
& Compositional correctness
& Yes \cmark
& No \xmark
& No \xmark
& No \xmark \\

\makecell[l]{HATIE\\}
& CVPR'25
& Editing
& Human-aligned editing quality
& Yes \cmark
& No \xmark
& No \xmark
& Limited \\

\makecell[l]{NTIRE 2025 T2I QA\\}
& CVPR'25
& T2I
& Fine-grained alignment and structural distortion
& Yes \cmark
& No \xmark
& Partial
& No \xmark \\

\makecell[l]{ICE-Bench\\}
& ICCV'25
& T2I + Editing
& Unified evaluation for creation and editing
& Yes \cmark
& No \xmark
& No \xmark
& No \xmark \\

\makecell[l]{Task-Decomposed AutoEval\\}
& ACL'25
& T2I evaluator
& Automatic multi-criterion quality assessment
& Yes \cmark
& No \xmark
& Partial
& Limited \\

\midrule
\textbf{HealBench (ours)}
& \textbf{AAAI'27}
& \textbf{T2I + Editing}
& \textbf{Perceived psychological comfort}
& \textbf{Yes} \cmark
& \textbf{Yes} \cmark
& \textbf{Yes} \cmark
& \textbf{Yes} \cmark \\
\bottomrule
\end{tabular}
}
\end{table*}

\begin{figure*}
  \centering
  \includegraphics[width=1\linewidth]{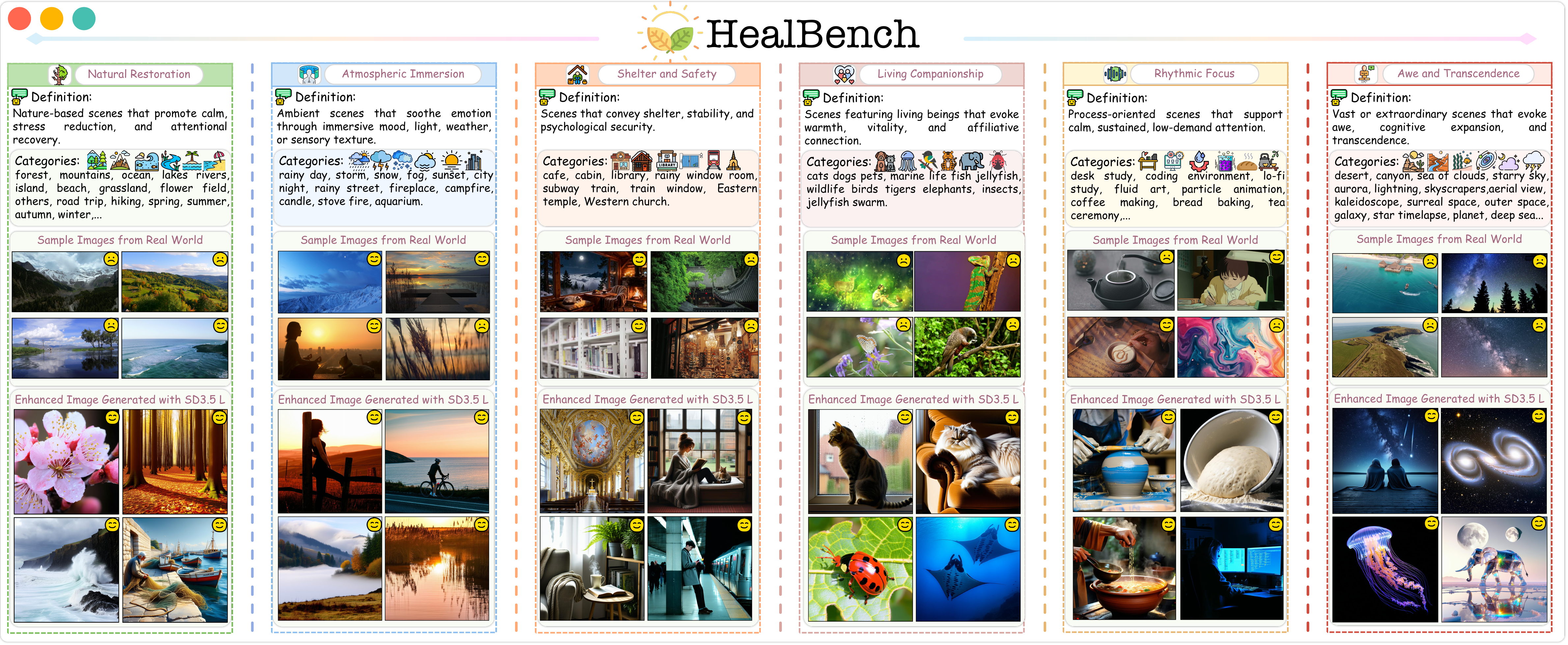}
  \vspace{-20pt}
  \caption{Overview of HealBench. HealBench organizes healing-related visual content into 19 categories and 92 subcategories, which are further grouped into six major classes according to different healing functions. For each category, we present both real-world examples and enhanced images generated via SD 3.5-L, which is also used for dataset augmentation.}
  \label{fig:begin}
  \vspace{-10pt}
\end{figure*}

We argue that this limitation is not merely a weakness of current models, but also a consequence of how the target has been formulated and evaluated. Existing generation pipelines are typically optimized for text-image alignment, perceptual quality, editing fidelity, or generic human preference. These objectives are valuable, but they do not directly capture the affective and restorative experience required by psychological comfort. In particular, current systems suffer from three mismatches. First, there is a \textbf{\textit{target mismatch}}: \textit{perceived psychological comfort} depends on affective and restorative cues that are not explicitly captured by existing optimization targets \cite{shi2025impact, belanche2025customer, ortega2025integration}. Second, there is an \textbf{\textit{evaluation-granularity mismatch}}: comfort-related failure may arise from different sources, including prompt misalignment, incorrect healing semantics, or poor image quality \cite{gao2025object, ortega2025integration}. As illustrated in Figure~\ref{fig:intro2}, two images may receive similarly low comfort ratings for fundamentally different reasons, making a single scalar score insufficient for diagnostic evaluation. Third, there is a \textbf{\textit{benchmark-and-feedback mismatch}}: as summarized in Table~\ref{tab:benchmark_comparison}, existing benchmarks and evaluators mainly target preference, alignment, fidelity, compositionality, or generic quality, but do not explicitly target \textit{perceived psychological comfort} or provide actionable, rationale-bearing feedback.
Against this background, we ask: \textcolor{DeepBlue}{how can perceived psychological comfort be turned into an evaluable and optimization-relevant objective for image generation and image editing?} Rather than treating comfort as a vague holistic preference, we formulate healing-oriented generation through three necessary dimensions: \textbf{text-image alignment}, \textbf{heal accuracy}, and \textbf{image quality}. A generated image should be considered healing only when it satisfies all three dimensions; failure in any one dimension is sufficient to disqualify the output. Accordingly, our goal is not merely to rank images from better to worse, but to identify whether, where, and why a generated image fails to provide perceived psychological comfort.

\vspace{-10pt}
Based on this formulation, we propose \textbf{HealBench} (shown in Figure \ref{fig:begin}), a fine-grained benchmark for evaluating the perceived psychological comfort of generated images. HealBench is constructed through a human-and-expert-in-the-loop pipeline that curates high-quality data from a large-scale candidate image-text pool, performs imitative data augmentation, and builds a rigorously filtered test set for unified comparison across models and prompting strategies. Unlike preference-style benchmarks that primarily produce aggregate scores, HealBench is designed around failure-based diagnosis: it identifies whether an output violates text-image alignment, heal accuracy, or image quality, thereby revealing different ways in which a generated image may fail to be healing.
Evaluation alone, however, is insufficient for improving generation systems. To make diagnosis actionable, we introduce a \textbf{score-plus-rationale} feedback paradigm, where each result is paired with both a comfort score and an explicit explanation of which dimension fails and why. This structured feedback supports prompt rewriting for both T2I generation and instruction-guided image editing. Since large-scale model iteration cannot rely entirely on costly human evaluation, we further train an automatic evaluator, Tri-dimensional Healing Scorer (\textbf{THS}), on top of HealBench to approximate these structured judgments and study its use as an optimization signal for downstream improvement, including RL-based fine-tuning.

Within this work, we treat psychological comfort as a distinct human-centered objective for visual generation, rather than as a by-product of semantic alignment, aesthetics, or generic preference. Our contributions are as follows:
(1)We formulate \textit{perceived psychological comfort} as a failure-based evaluation problem for healing-oriented visual generation, defined by three necessary dimensions: \textbf{text-image alignment}, \textbf{heal accuracy}, and \textbf{image quality}.
(2)We build \textbf{HealBench}, a fine-grained benchmark that diagnoses whether a generated image fails on any of the three dimensions, instead of relying solely on aggregate preference-style scoring.
(3)We introduce a \textbf{score-plus-rationale} feedback paradigm and \textbf{THS}, turning structured diagnosis into an actionable signal for prompt refinement and downstream model optimization.

\section{Related Work}

\noindent\textbf{Image Generation and Editing Models.}
Recent visual generation models have made substantial progress in fidelity, semantic alignment, controllability, and human preference alignment. Diffusion models such as DDPM, latent diffusion, Imagen, and DiT enable high-resolution synthesis and strong prompt following \cite{ho2020denoising, rombach2022high, saharia2022photorealistic, peebles2023scalable}; autoregressive and token-based models such as DALL$\cdot$E, Make-A-Scene, and Muse improve complex content modeling and composition \cite{ramesh2021zero, gafni2022make, chang2023muse}; and flow-matching or rectified-flow methods further enhance training and sampling efficiency \cite{lipmanflow, liu2025flashaudio, esser2024scaling}. Preference optimization and reinforcement learning methods further align T2I models with human judgments \cite{blacktraining, fan2023dpok, wallace2024diffusion}, while text-guided editing methods provide instruction-based control over image content and style \cite{hertzprompt, mokady2023null, brooks2023instructpix2pix}. However, these methods mainly optimize whether images are faithful, realistic, editable, or generally preferred. Existing affective methods, such as Affective Image Filter and EmoEdit, focus on explicit emotion cues or targeted emotional editing \cite{weng2023affective, yang2025emoedit}, leaving subtler outcomes such as psychological comfort, safety, companionship, and healing underexplored.

\noindent\textbf{Text-to-Image Evaluation Benchmarks and Metrics.}
Existing T2I evaluation primarily targets semantic correctness, visual quality, compositional reasoning, and generic preference. CLIP-based metrics estimate text-image alignment \cite{radford2021learning}; reward and preference models such as PickScore, ImageReward, HPS, and HPSv3 capture overall human preference \cite{kirstain2023pick, xu2023imagereward, wu2023human, ma2025hpsv3}; and aesthetic metrics such as AVA and NIMA assess visual appeal \cite{murray2012ava, talebi2018nima}. Broader benchmarks, including DrawBench, GenEval, T2I-CompBench, VQAScore / GenAI-Bench, and HEIM, evaluate compositionality, spatial relations, reasoning, robustness, safety, and overall quality \cite{saharia2022photorealistic, ghosh2023geneval, huang2023t2i, lee2023holistic, lin2024evaluating, li2024evaluating}. Yet these metrics mainly ask whether an image is correct, high-quality, or preferred, rather than whether it is psychologically comforting. Although HRS-Bench introduces emotion-grounded evaluation \cite{bakr2023hrs}, and HEIM and GenAI-Bench cover broad evaluation dimensions \cite{lee2023holistic, li2024evaluating}, deeper affective outcomes such as soothingness and emotional comfort remain insufficiently evaluated. Recent studies further show that metric rankings can be unstable across prompts and protocols \cite{wilesrevisiting}, and that faithfulness metrics remain unreliable on challenging subsets \cite{saxon2024evaluates}. HealBench addresses this gap by treating perceived psychological comfort as a dedicated, fine-grained, and failure-diagnostic evaluation target.

\begin{figure*}[t]
    \centering
    \includegraphics[width=1\textwidth]{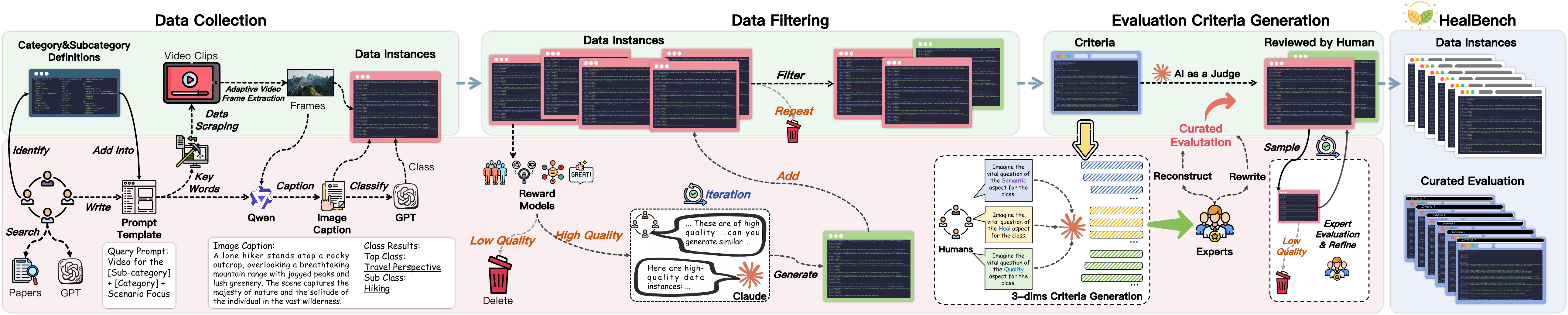}
    \caption{Overview of the HealBench construction pipeline. The benchmark is built through three stages: data collection, reward-model-based filtering and augmentation, and tri-dimensional evaluation-criteria generation. Human and expert review is incorporated throughout the pipeline to ensure the quality of both data instances and evaluation criteria.}
    \label{fig:healbench}
    \vspace{-8pt}
\end{figure*}
\begin{figure*}[t]
    \centering
    \includegraphics[width=1\textwidth]{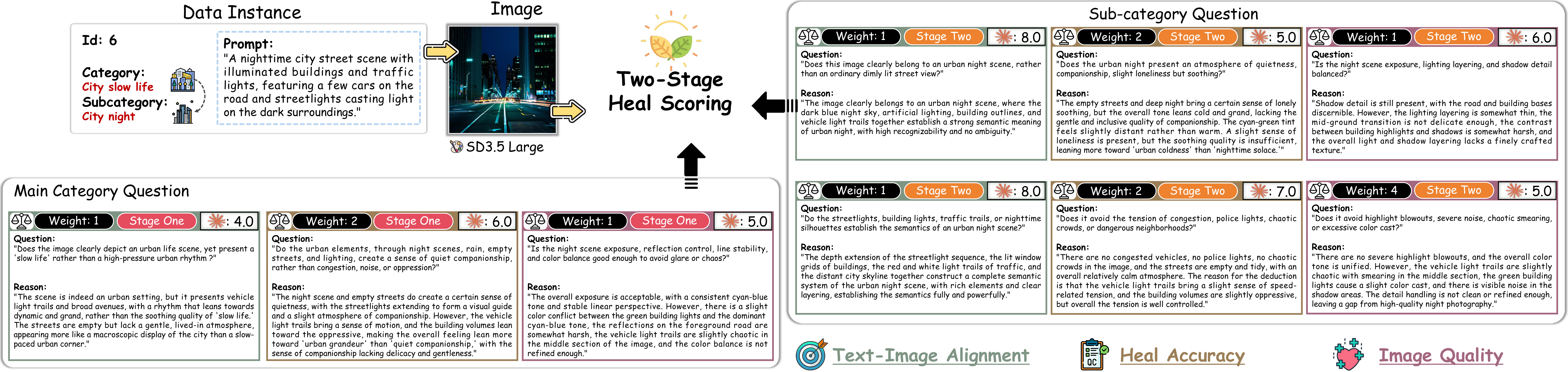}
    \caption{Overview of the two-stage HealScore pipeline. HealScore first evaluates each image with three category-level questions on \textit{text-image alignment}, \textit{heal accuracy}, and \textit{image quality}, and then refines the judgment with six subcategory-level questions. The final score is computed by weighted aggregation of the resulting multi-dimensional responses.}
    \label{fig:healscore}
    \vspace{-15pt}
\end{figure*}

\section{Method}

\subsection{Overview}
\textbf{HealBench} is constructed in four stages:
(1) taxonomy design and web-video-based image collection;
(2) captioning, hierarchical labeling, and reward-model-based filtering;
(3) category-aware evaluation protocol construction and HealScore annotation;
(4) tri-dimensional scorer training and data augmentation.
Experts participate in taxonomy design, question revision, and iterative quality checks, while VLMs/LLMs are used as scalable assistants.
\vspace{-8pt}

\subsection{Data Collection and Filtering}
We first build a two-level taxonomy for psychologically comforting imagery. Experts define 18 coarse-grained categories, supplemented with an \textit{Others} class, to cover common restorative visual functions. The coarse-grained category set is
$\mathcal{T}=\{T_i\}_{i=1}^{19}.$
For each coarse category \(T_i\), we further define a fine-grained subcategory set
$\mathcal{S}_i=\{S_{i,j}\}_{j=1}^{n_i},$
which also includes an \textit{Others} subtype. In total, the taxonomy contains 19 coarse-grained categories and 102 fine-grained subcategories.
Each image \(I\) is assigned to exactly one category pair:
\begin{equation}
c(I)=(T_i,S_{i,j}), \quad T_i\in\mathcal{T},\; S_{i,j}\in\mathcal{S}_i.
\end{equation}
Based on this taxonomy, we manually design category-aware retrieval templates to crawl web videos and obtain a raw video set ($\mathcal{D}_{\text{video}}$). We then extract frames from these videos, producing a raw image set ($\mathcal{D}_{\text{raw}}$) of about 100K frames. For each image in $\mathcal{D}_{\text{raw}}$, we use \textsc{Qwen2.5-VL-7B-Instruct} to generate a caption, and then use Claude to assign hierarchical labels. Manual inspection removes invalid or ambiguous category pairs, leaving 92 valid pairs for subsequent experiments. 
The category pairs and the detailed data collection and annotation process are shown in Fig.\ref{fig:healbench}.
To remove noisy samples caused by semantic drift, poor composition, and low visual quality, we rank images with four complementary reward signals: CLIP Score, Aesthetic Score, PickScore, and HPSv3. These models serve as coarse quality gates rather than final healing labels. Denote their scores by \(r_{\text{clip}}, r_{\text{aes}}, r_{\text{pick}}, r_{\text{hps}}\), respectively. After independent normalization, we compute the overall score as
\begin{equation}
r_{\text{overall}}
=
\tilde{r}_{\text{clip}}
+
\tilde{r}_{\text{aes}}
+
\tilde{r}_{\text{pick}}
+
\tilde{r}_{\text{hps}}.
\end{equation}
Within each fine-grained subcategory, we rank images by \(r_{\text{overall}}\) and retain the top 33\%, which removes low-quality candidates while preserving category diversity, yielding the filtered real-image subset \(\mathcal{D}_{\text{filter}}\).
\vspace{-8pt}

\subsection{Evaluation Criteria and HealScore}
Existing image-generation metrics mainly emphasize semantic alignment and aesthetic preference, but are insufficient for evaluating psychological soothing effects. We therefore propose \textit{HealScore}, a structured metric defined on three dimensions: text-image alignment, heal accuracy, and image quality. Here, heal accuracy refers to the perceived healing degree of the image under its target category, not to clinical efficacy.
For each category pair \(c=(T_i,S_{i,j})\), we build a category-aware question set. An LLM first generates one coarse-level question for the general healing function of \(T_i\) and two fine-grained questions for the concrete visual or affective cues of \(S_{i,j}\) under each dimension. Experts then revise the questions to remove ambiguous, redundant, or category-mismatched formulations. The resulting question set is
\begin{equation}
\mathcal{Q}_c=
\left\{
q^{c}_{d,k}
\mid
d\in\{\text{sem},\text{heal},\text{qual}\},\;
k\in\{0,1,2\}
\right\},
\end{equation}
where \(q^{c}_{d,0}\) is the coarse-level question and \(q^{c}_{d,1}, q^{c}_{d,2}\) are fine-grained supplementary questions. 
For each image \(I\in\mathcal{D}_{\text{filter}}\), an LLM scores every question in \(\mathcal{Q}_{c(I)}\) on a 1--10 scale. Let the resulting scores be \(r_{\text{sem},k}(I)\), \(r_{\text{heal},k}(I)\), and \(r_{\text{qual},k}(I)\). We aggregate them within each dimension:
\begin{equation}
s_{d}(I)=\sum_{k=0}^{2} r_{d,k}(I), 
\quad d\in\{\text{sem},\text{heal},\text{qual}\}.
\end{equation}
We report the three dimension scores for diagnosis and use a compact summary score only for ranking, training, and large-scale comparison. Since psychological comfort is the primary objective of the benchmark, the summary score gives the healing dimension twice the weight of the alignment and quality dimensions:
\begin{equation}
\text{HealScore}(I)
=
s_{\text{sem}}(I)
+
2\,s_{\text{heal}}(I)
+
s_{\text{qual}}(I).
\end{equation}
The 1:2:1 design therefore emphasizes the target dimension while retaining the two supporting constraints, rather than masking a failed dimension in diagnostic interpretation. To reduce annotation noise, we manually inspect sampled scored results and iteratively refine the question set by expert feedback. The detailed scoring process of HealScore is shown in Fig.\ref{fig:healscore}.

To further enrich the dataset, we select, for each category pair, the top-10 images ranked by HealScore and the top-10 images ranked by the normalized reward-model total score. Their captions are rewritten by Claude and used as prompts for \textsc{SD3.5-Large} to synthesize additional images, producing an augmented set \(\mathcal{D}_{\text{augment}}\). We merge \(\mathcal{D}_{\text{augment}}\) with \(\mathcal{D}_{\text{filter}}\) and remove near-duplicates based on image similarity, obtaining the final dataset
\(\mathcal{D}_{\text{heal}}\).

\vspace{-8pt}
\subsection{Tri-dimensional Heal Scorer (THS)}
To approximate rule-based HealScore with a learnable model, we train a \textit{Tri-dimensional Heal Scorer (THS)}. Given an image and its category pair \((T_i,S_{i,j})\), THS predicts three dimension-specific scores for text-image alignment, heal accuracy, and image quality, and combines them into the summary score.
We adopt BLIP image-text matching as the backbone. Given image \(I\) and category text
\begin{equation}
x=\texttt{"An image of } T_i \texttt{, } S_{i,j} \texttt{"},
\end{equation}
we encode image and text separately, project them into a shared embedding space, and obtain normalized embeddings \(v\) and \(t\). Their fused representation is defined as
$h=[v; t; v\odot t],$
where \([\,\cdot\,;\,\cdot\,]\) denotes concatenation and \(\odot\) denotes element-wise multiplication.
The fused feature is fed into three independent MLP heads to predict the three dimension scores.
Let the supervision targets be
$y_{\text{sem}},y_{\text{heal}}, \text{and} \, y_{\text{qual}},$
with corresponding predictions
$
\hat{y}_{\text{sem}},\hat{y}_{\text{heal}}, \text{and} \, \hat{y}_{\text{qual}}.
$
We first optimize the dimension-wise regression loss:
\begin{equation}
\mathcal{L}_{\text{dim}}
=
\sum_{d\in\{\text{sem},\text{heal},\text{qual}\}}
\left\|\hat{y}_d-y_d\right\|^2.
\end{equation}
To enforce consistency with the summary HealScore while preserving dimension-wise supervision, we further define
\begin{equation}
\begin{aligned}
y_{\text{total}}&=y_{\text{sem}}+2y_{\text{heal}}+y_{\text{qual}},\\
\hat{y}_{\text{total}}&=\hat{y}_{\text{sem}}+2\hat{y}_{\text{heal}}+\hat{y}_{\text{qual}},
\end{aligned}
\end{equation}
and apply an additional total-score loss:
$
\mathcal{L}_{\text{total}}
=
\left\|\hat{y}_{\text{total}}-y_{\text{total}}\right\|^2.
$
The final objective is
$
\mathcal{L}
=
\mathcal{L}_{\text{dim}}
+
\lambda \mathcal{L}_{\text{total}},
$
where \(\lambda\) balances the two terms.
We optimize THS with AdamW, using different learning rates for the pretrained backbone and the newly initialized heads. During training, we retain only samples with complete annotations on all three dimensions, and select the final model according to the validation error on \(\mathcal{L}_{\text{total}}\).

\section{Experiments}
\label{sec:exp}

\subsection{Overview}
We evaluate HealBench from three perspectives: validity, utility, and scalability, covering benchmark representativeness, tri-dimensional diagnosis, feedback-based improvement, automatic scoring, and downstream optimization.
\vspace{-8pt}

\begin{figure}[htbp]
    \centering
    \includegraphics[width=1\linewidth]{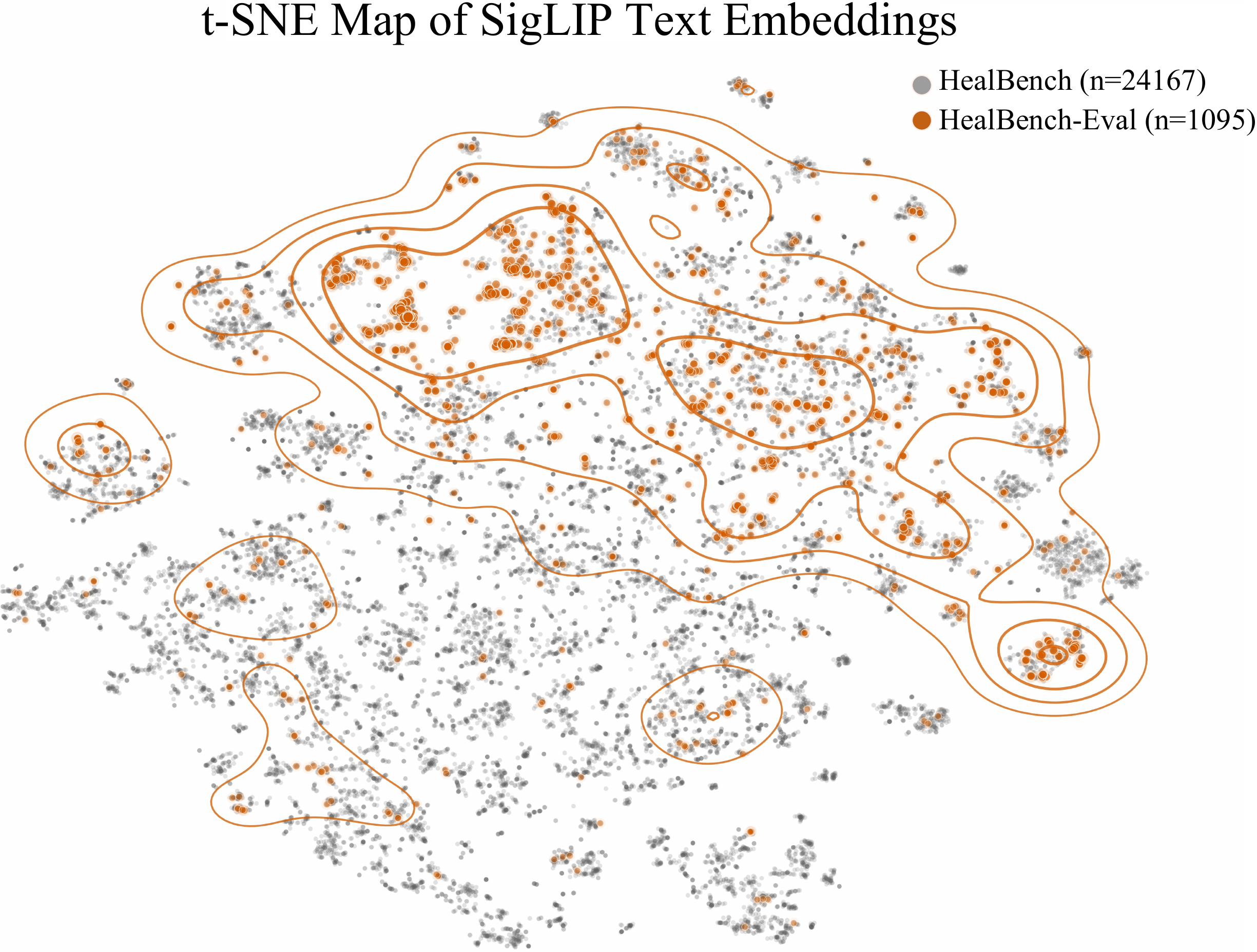}
    \caption{t-SNE visualization of SigLIP embeddings for the full HealBench data and the selected HealBench-Eval. 
}
    \label{fig:tsne}
\end{figure}

\begin{table}[htbp]
\centering
\caption{Agreement between HealScore-induced rankings and pairwise preferences from human annotators and VLM-based judges.}\vspace{-8pt}
\label{tab:pref_compare}
\resizebox{\columnwidth}{!}{%
\begin{tabular}{lccc}
\toprule
Model & Accuracy & Kendall's $\tau$ & Spearman's $\rho$ \\
\midrule
Haiku  & 0.6831 & 0.3546 & 0.4319 \\ 
Sonnet 4-6 & 0.6933 & 0.3868 & 0.4711 \\
People & 0.8012 & 0.4487 & 0.5448 \\
\bottomrule
\end{tabular}\vspace{-15pt}
}
\end{table}

\subsection{HealBench-Eval Construction and Validation}
\noindent\greenbudge{\textbf{Construction}}
To enable controlled evaluation without sacrificing diversity, we construct a compact HealBench-Eval from the filtered real-image subset. For each of the 92 valid category pairs, we extract SigLIP embeddings, run K-means clustering, and select representative samples. Figure~\ref{fig:tsne} shows that the resulting subset follows the overall HealBench distribution rather than concentrating in a few dense regions, making it suitable for controlled healing-oriented evaluation.

\noindent\greenbudge{\textbf{Validation}}To test whether the tri-dimensional formulation captures a meaningful healing-oriented signal, we compare HealScore-induced rankings with human and VLM preferences on 1,500 image pairs. Annotators, Claude-Haiku, and Claude-Sonnet 4-6 select the more psychologically comforting image. Table~\ref{tab:pref_compare} shows that human judgments achieve the highest agreement, while Sonnet slightly outperforms Haiku, indicating that HealScore provides a meaningful approximation of this structured preference.

\begin{table}[htbp]
\centering
\caption{Quantitative comparison of different text-to-image models across multiple evaluation metrics.}\vspace{-8pt}
\label{tab:t2i_compare_wide}
\scriptsize
\setlength{\tabcolsep}{4pt}
\renewcommand{\arraystretch}{1.05}
\resizebox{\columnwidth}{!}{%
\begin{tabular}{lcccc}
\toprule
Model & LAION Aesthetic $\uparrow$ & CLIPScore $\uparrow$ & PickScore $\uparrow$ & HPSv3 $\uparrow$ \\
\midrule
SD 2.1            & 4.403 & 0.885 & 0.785 & 3.406 \\
SDXL Base 1.0     & 4.631 & 0.885 & 0.790 & 3.602 \\
pixArt-XL    & 4.717 & 0.946 & 0.864 & 12.623 \\
SDXL Turbo     & 4.095 & 0.975 & 0.812 & 5.543 \\
Playground v2.5 & 5.164 & \textbf{0.988} & \textbf{0.899} & 12.712 \\
Hunyuan-Dit         & \textbf{5.271} & 0.976 & 0.872 & 10.991 \\
SD 3 Medium          & 4.841 & 0.965 & 0.863 & 11.515 \\
Kolors          & 4.638 & 0.901 & 0.823 & 9.587 \\
FLUX.1 [dev]      & 4.798 & 0.938 & 0.876 & \textbf{13.585} \\
FLUX.1 [schnell]  & 4.682 & 0.943 & 0.840 & 10.624 \\
SD 3.5 Large         & 4.912 & 0.971 & 0.867 & 11.792 \\
SD 3.5 Medium         & 4.867 & 0.974 & 0.867 & 11.533 \\
sana16          & 4.666 & 0.967 & 0.870 & 12.955 \\
Lumina-Image 2.0  & 4.950 & 0.984 & 0.864 & 11.919 \\
\midrule
GT              & 4.954 & 0.925 & 0.838 & 9.797 \\
\bottomrule
\end{tabular}\vspace{-15pt}
}
\end{table}

\newcommand{\upcell}[2]{%
\makecell[c]{#1\\[-0.25em]\scriptsize\textcolor[rgb]{0.85,0.33,0.10}{$\uparrow$#2}}%
}

\newcommand{\blueupcell}[2]{%
\makecell[c]{#1\\[-0.25em]\scriptsize\textcolor[rgb]{0.00,0.45,0.74}{$\uparrow$#2}}%
}

\begin{table*}[htbp]
\centering
\scriptsize
\setlength{\tabcolsep}{4pt}
\caption{Baseline summary table. All metrics are positive indicators. For prompt-rewrite models, the orange mark denotes the improvement over the corresponding non-rewrite base model. For image editing models, the blue mark denotes the improvement over the corresponding original model.}\vspace{-8pt}
\label{tab:baseline_summary}
\resizebox{\textwidth}{!}{%
\begin{tabular}{lcccccccccc}
\toprule
Model & Overall & Text-Image Alignment & Heal Accuracy & Image Quality & Natural Restoration & Atmospheric Immersion & Shelter and Safety & Living Companionship & Rhythmic Focus & Awe and Transcendence \\
\midrule

\multicolumn{11}{c}{\textit{Diffusion Models}} \\
\midrule
SD 2.1 & 19.390 & 5.172 & 4.820 & 4.578 & 18.754 & 19.678 & 18.484 & 21.780 & 19.412 & 20.147 \\
SDXL Base 1.0 & 20.723 & 5.319 & 5.173 & 5.057 & 20.258 & 21.269 & 21.915 & 24.284 & 19.859 & 20.481 \\
PixArt-XL & 25.080 & 6.435 & 6.169 & 6.308 & 24.953 & 24.832 & 25.115 & 27.787 & 25.036 & 24.844 \\
SDXL Turbo & 18.765 & 5.419 & 4.517 & 4.312 & 17.372 & 18.689 & 17.536 & 19.837 & 20.549 & 18.695 \\
Playground v2.5 & 25.088 & 6.479 & 6.094 & 6.422 & 24.551 & 25.304 & 24.449 & 28.177 & 25.162 & 25.253 \\
Hunyuan-DiT & 25.020 & 6.378 & 6.178 & 6.286 & 24.341 & 24.849 & 24.286 & 28.057 & 25.427 & 25.347 \\
Kolors & 23.535 & 6.005 & 5.866 & 5.797 & 23.586 & 23.924 & 22.832 & 26.369 & 23.267 & 23.255 \\
Sana-1.6B & 24.419 & 6.386 & 5.917 & 6.201 & 23.748 & 25.044 & 24.291 & 29.624 & 23.821 & 24.541 \\

\midrule
\multicolumn{11}{c}{\textit{Flow Matching Models}} \\
\midrule
SD 3 Medium & 23.465 & 6.121 & 5.731 & 5.882 & 22.556 & 23.691 & 21.617 & 28.560 & 23.462 & 24.780 \\
FLUX.1 [dev] & 24.836 & 6.289 & 6.152 & 6.243 & 24.211 & 24.195 & 23.489 & 29.759 & 25.424 & 25.222 \\
FLUX.1 [schnell] & 22.717 & 5.954 & 5.561 & 5.642 & 22.080 & 21.632 & 21.108 & 27.830 & 23.225 & 23.913 \\
SD 3.5 Large & 24.571 & 6.346 & 6.008 & 6.210 & 23.694 & 24.381 & 23.160 & 29.057 & 25.106 & 25.154 \\
SD 3.5 Medium & 24.164 & 6.321 & 5.884 & 6.075 & 23.285 & 24.131 & 22.391 & 28.730 & 24.266 & 25.620 \\
Lumina-Image 2.0 & 23.169 & 6.132 & 5.627 & 5.782 & 22.483 & 22.855 & 22.035 & 28.447 & 23.433 & 23.602 \\

\midrule
\multicolumn{11}{c}{\textit{Prompt-Rewrite Models}} \\
\midrule
Hunyuan-DiT (Rewrite)
& \upcell{26.631}{1.612}
& \upcell{6.617}{0.239}
& \upcell{6.711}{0.533}
& \upcell{6.592}{0.306}
& \upcell{26.841}{2.500}
& \upcell{25.843}{0.994}
& \upcell{26.336}{2.050}
& \upcell{29.546}{1.489}
& \upcell{27.183}{1.757}
& \upcell{25.574}{0.226} \\
FLUX.1 [schnell] (Rewrite)
& \upcell{24.514}{1.797}
& \upcell{6.328}{0.374}
& \upcell{6.051}{0.490}
& \upcell{6.084}{0.442}
& \upcell{24.411}{2.331}
& \upcell{23.090}{1.459}
& \upcell{23.942}{2.835}
& \upcell{28.794}{0.965}
& \upcell{25.043}{1.818}
& \upcell{24.552}{0.638} \\
SD 3.5 Large (Rewrite)
& \upcell{26.149}{1.578}
& \upcell{6.638}{0.293}
& \upcell{6.471}{0.464}
& \upcell{6.568}{0.358}
& \upcell{25.851}{2.157}
& \upcell{24.993}{0.611}
& \upcell{25.825}{2.664}
& \upcell{31.057}{2.000}
& \upcell{26.680}{1.575}
& \upcell{25.836}{0.683} \\
SD 3.5 Medium (Rewrite)
& \upcell{25.598}{1.434}
& \upcell{6.556}{0.235}
& \upcell{6.322}{0.438}
& \upcell{6.398}{0.323}
& \upcell{25.071}{1.786}
& \upcell{24.683}{0.552}
& \upcell{24.554}{2.163}
& \upcell{29.979}{1.248}
& \upcell{26.179}{1.912}
& \upcell{26.046}{0.426} \\

\midrule
\multicolumn{11}{c}{\textit{Image Editing Models}} \\
\midrule
Pix2Pix (Original) & 20.666 & 5.454 & 5.081 & 5.050 & 20.349 & 21.401 & 20.832 & 23.837 & 20.832 & 19.059 \\
Pix2Pix (Rewrite)
& \blueupcell{22.802}{2.136}
& \blueupcell{5.853}{0.399}
& \blueupcell{5.664}{0.583}
& \blueupcell{5.620}{0.570}
& \blueupcell{23.047}{2.698}
& \blueupcell{23.193}{1.792}
& \blueupcell{23.649}{2.817}
& \blueupcell{25.674}{1.837}
& \blueupcell{22.719}{1.887}
& \blueupcell{20.606}{1.547} \\
SDXL-Pix2Pix (Original) & 19.307 & 5.137 & 4.722 & 4.727 & 18.612 & 19.576 & 18.351 & 23.780 & 19.272 & 19.733 \\
SDXL-Pix2Pix (Rewrite)
& \blueupcell{22.053}{2.746}
& \blueupcell{5.711}{0.574}
& \blueupcell{5.445}{0.723}
& \blueupcell{5.453}{0.726}
& \blueupcell{22.111}{3.499}
& \blueupcell{22.575}{2.999}
& \blueupcell{21.464}{3.113}
& \blueupcell{26.255}{2.475}
& \blueupcell{21.908}{2.636}
& \blueupcell{20.917}{1.184} \\

\midrule
\multicolumn{11}{c}{\textit{Ground Truth}} \\
\midrule
Real Images & 24.528 & 6.375 & 5.995 & 6.163 & 23.186 & 24.119 & 25.205 & 27.901 & 25.216 & 24.503 \\
\bottomrule
\end{tabular}%
}
\vspace{-10pt}
\end{table*}

\vspace{-8pt}
\subsection{Benchmarking Baseline T2I Models}

We next benchmark representative diffusion and flow-matching T2I models under the original prompts on healing-oriented generation.
Table~\ref{tab:t2i_compare_wide} reports standard metrics as reference baselines. Recent models perform strongly on Aesthetic score, CLIPScore, PickScore, and HPSv3, sometimes exceeding the real-image reference on individual metrics. Because these metrics do not target perceived psychological comfort, they should not be treated as evidence of healing-oriented generation quality.

\noindent\textbf{Overall healing-oriented performance.}
Table~\ref{tab:baseline_summary} shows clear gaps across model families. Among original baselines, \textit{Playground v2.5} (25.088), \textit{PixArt-XL} (25.080), and \textit{Hunyuan-DiT} (25.020) perform best overall, while \textit{SDXL Turbo} (18.765) and \textit{SD 2.1} (19.390) lag behind. Even when advanced generators approach the real-image reference, their scores remain uneven across dimensions and healing functions.

\noindent\textbf{Tri-dimensional analysis.}
In Table \ref{tab:baseline_summary}, the three dimensions reveal why aggregate scores are insufficient. Strong baselines usually score above 6.0 on \textit{Text-Image Alignment} and \textit{Image Quality}, but are lower and more variable on \textit{Heal Accuracy}: \textit{Playground v2.5} obtains 6.479/6.422 on Alignment/Quality but 6.094 on Heal Accuracy, and \textit{SD 3.5 Medium} obtains 6.321/6.075 but 5.884. Thus, perceived healing degree cannot be inferred from alignment or visual quality alone.

\noindent\textbf{Class-level analysis.}
As shown in Table~\ref{tab:baseline_summary}, Class-wise results further show uneven difficulty. \textit{Living Companionship} is often easiest, reaching 29.624 for \textit{Sana-1.6B}, 29.759 for \textit{FLUX.1 [dev]}, and 29.057 for \textit{SD 3.5 Large}. In contrast, \textit{Shelter and Safety} and \textit{Natural Restoration} show larger model-dependent variation, and no model dominates all six classes. HealBench therefore exposes both dimension-specific and function-specific limitations.
\vspace{-8pt}

\begin{figure}[htbp]
    \centering
    \includegraphics[width=1.0\linewidth]{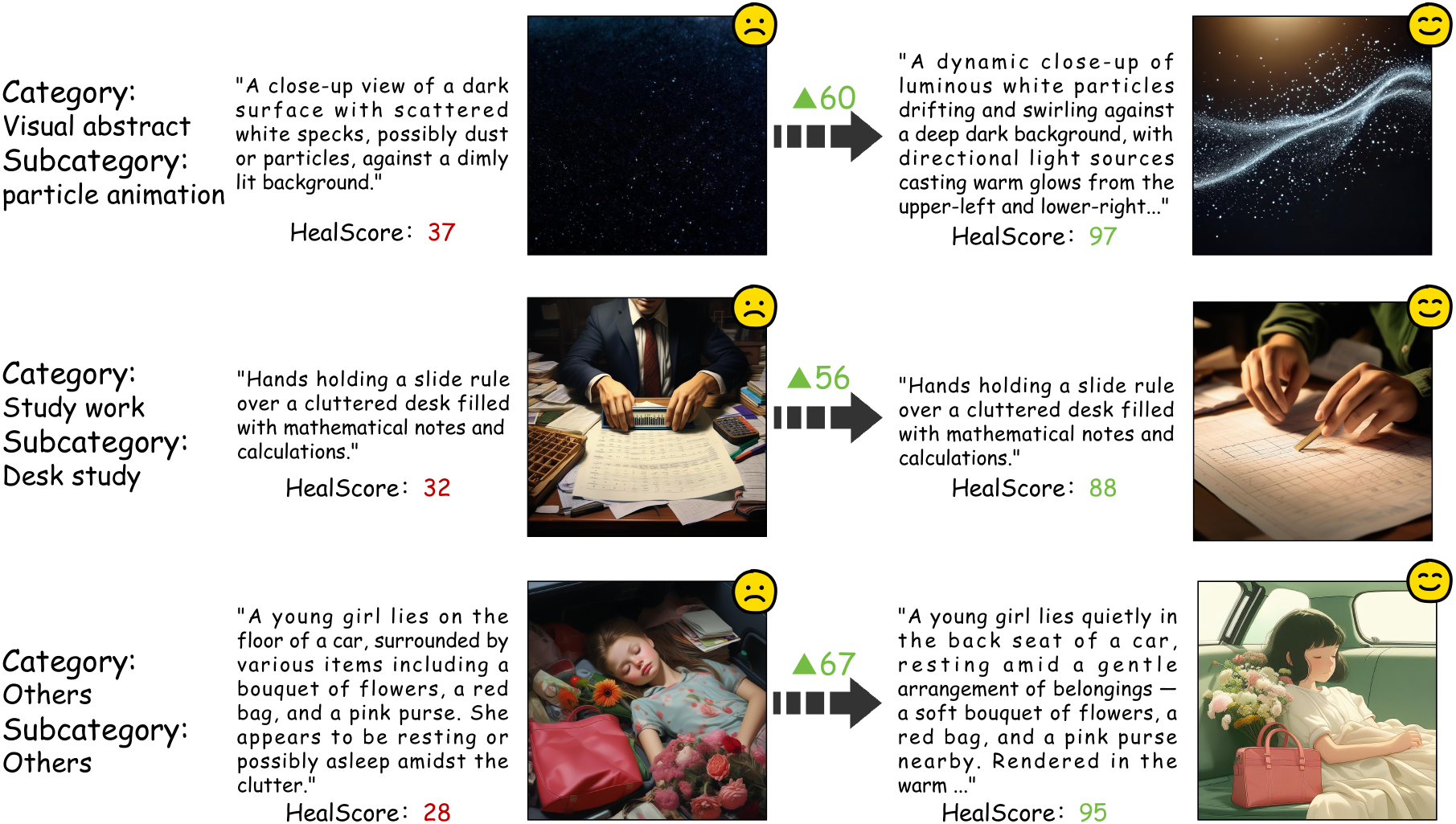} \vspace{-8pt}
    \caption{Qualitative examples of our prompt rewriting strategy. Using the same image generation model, targeted prompt rewriting markedly improves generation outcomes, yielding more coherent and safer images with substantially higher HealScores across multiple categories.} 
    \vspace{-10pt}
    \label{fig:re_image}      
\end{figure}

\subsection{Prompt Rewriting as Intervention}
We next test whether score-plus-rationale feedback can improve prompts on healing-oriented generation for both text-to-image generation and image editing.

\noindent\bluebadge{\textbf{Prompt rewriting improves healing-oriented generation.}}
Table~\ref{tab:baseline_summary} shows that rewriting improves all tested T2I models, including \textit{Hunyuan-DiT}, \textit{FLUX.1 [schnell]}, \textit{SD 3.5 Large}, and \textit{SD 3.5 Medium}, with the most consistent gains on \textit{Heal Accuracy}.

\noindent\bluebadge{\textbf{Prompt rewriting also benefits image editing.}}
For editing, rewritten prompts improve both \textit{Pix2Pix} and \textit{SDXL-Pix2Pix} in overall HealScore and all three dimensions.

\noindent\textbf{Qualitative analysis.}
Figure~\ref{fig:re_image} shows that rewritten prompts produce more coherent, less visually tense images aligned with the intended healing semantics.

\begin{table}[htbp]
\centering
\small
\setlength{\tabcolsep}{4pt}
\renewcommand{\arraystretch}{1.1}
\caption{User study on personalized healing delivery with 24 subthreshold-depression college students. 
Participants in the personalized condition could choose among the six HEAL classes, while those in the random condition viewed randomly assigned healing images. 
Mood change was measured with HAMD-17 after one hour of viewing. }\vspace{-8pt}
\label{tab:user_study_delivery}
\resizebox{\columnwidth}{!}{%
\begin{tabular}{lccc}
\toprule
Condition & $N$ & Improved Ratio (\%) $\uparrow$ & HAMD-17 $\downarrow$ \\
\midrule
Random playback & 12 & 45.5 & -1.27 \\
Personalized class selection & 12 & 75.0 & -3.42 \\
\midrule
Difference / Gain & -- & +29.5 & -2.15 \\
\bottomrule
\end{tabular}
}
\vspace{-10pt}
\end{table}

\subsection{User Study on Personalized Healing Delivery}

In a preliminary study with 24 college students with subthreshold depression, participants either choose a HEAL class or view randomly assigned healing images. Personalized selection yields a higher improvement ratio (75.0\% vs. 45.5\%) and larger HAMD-17 reduction (-3.42 vs. -1.27), suggesting that the six-class organization is useful beyond benchmark analysis.

\begin{table}[htbp]
\centering 
\scriptsize
\setlength{\tabcolsep}{2.5pt}
\renewcommand{\arraystretch}{0.95}
\caption{THS vs. generic scorers on pairwise healing-preference ranking.}\vspace{-8pt}
\resizebox{\columnwidth}{!}{%
\begin{tabular}{lccc}
\toprule 
Model & Acc. $\uparrow$ & Kendall's $\tau$ $\uparrow$ & Spearman's $\rho$ $\uparrow$ \\
\midrule
PickScore      & 0.5593          & 0.1525          & 0.2254          \\
LAION Aesthetic      & 0.5507          & 0.1077          & 0.1595          \\
CLIPScore      & 0.5413          & 0.1272          & 0.1882          \\
HPSv3          & 0.5013          & 0.0379          & 0.0595          \\
THS  & \textbf{0.8527} & \textbf{0.7741} & \textbf{0.9223} \\
\bottomrule
\end{tabular}
}
\label{tab:rm_compare}
\vspace{-8pt}
\end{table}

\vspace{-5pt}
\subsection{THS as an Automatic Healing Scorer}

We next examine whether THS can provide a more suitable automatic evaluation signal for healing-oriented generation than generic scoring models. 
To this end, we compare THS with several widely used baselines, including PickScore, LAION Aesthetic, CLIPScore, and HPSv3, using the same pairwise ranking protocol as in the preference validation experiment.
As shown in Table~\ref{tab:rm_compare}, THS substantially outperforms all generic baselines across all three ranking metrics, achieving 0.8527 in Accuracy, 0.7741 in Kendall's $\tau$, and 0.9223 in Spearman's $\rho$. 
By contrast, the generic scorers remain far behind, with Accuracy values clustered around 0.50--0.56 and much weaker rank correlation. 
These results indicate that existing scoring models for semantics, aesthetics, or broad human preference do not adequately capture the specific notion of perceived psychological comfort targeted in this work.
Results justify the need for a task-specific automatic scorer. 
Rather than serving only as a descriptive metric, THS provides a scalable evaluation signal that is better aligned with healing-oriented preference and can therefore support both large-scale benchmarking and downstream optimization.

\vspace{-5pt}
\begin{table}[htbp]
\centering
\caption{Main results of the base SD 3.5 Medium model and its Flow-GRPO fine-tuned variant. Higher values are highlighted in blue.}\vspace{-8pt}
\label{tab:sd35_flowgrpo_main}
\setlength{\tabcolsep}{3.2pt}
\renewcommand{\arraystretch}{1.15}
\small
\begin{tabular}{lcccc}
\toprule
Model & Overall & Semantic & Heal & Quality \\
\midrule
SD 3.5 Medi (base)
& 24.16
& 6.32
& 5.88
& 6.08 \\
SD 3.5 Medi + Flow-GRPO
& \textbf{25.67}
& \textbf{6.48}
& \textbf{6.45}
& \textbf{6.28} \\
\bottomrule
\end{tabular}
\vspace{-10pt}
\end{table}

\vspace{-5pt}
\subsection{Exploratory RL Fine-Tuning with THS Reward}

Finally, we conduct a preliminary study to examine whether THS can serve not only as an evaluator, but also as a reward signal for direct model optimization. 
To this end, we fine-tune \textit{SD 3.5 Medium} with Flow-GRPO using a reward defined by a weighted combination of THS and CLIPScore.
As shown in Table~\ref{tab:sd35_flowgrpo_main}, the Flow-GRPO fine-tuned model improves over the base \textit{SD 3.5 Medium} across all four metrics, including \textit{Overall} (23.47 $\rightarrow$ 25.67), \textit{Semantic} (6.12 $\rightarrow$ 6.48), \textit{Heal} (5.73 $\rightarrow$ 6.45), and \textit{Quality} (5.88 $\rightarrow$ 6.28). 
Notably, the largest gain appears on the \textit{Heal} dimension, which is consistent with the role of THS as a healing-oriented reward signal. 
This suggests that the proposed scorer is not only useful for benchmark evaluation, but can also provide an effective optimization signal for improving healing-oriented generation.
We emphasize that this experiment is exploratory rather than a fully optimized RL recipe. 
Nevertheless, the consistent gains across all dimensions provide initial evidence that THS can support downstream alignment toward psychologically comforting image generation.
\section{Conclusion}
\label{sec:dis}

We present \textbf{HealBench}, the first fine-grained benchmark for evaluating perceived psychological comfort in text-to-image generation and image editing. By explicitly defining healing-oriented generation along three complementary dimensions (\textit{text-image alignment}, \textit{heal accuracy}, and \textit{image quality}), we move psychological comfort from an implicit notion of preference to a structured and measurable objective. Building on this formulation, we develop a large-scale human-and-expert-in-the-loop benchmark, introduce the rule-based \textbf{HealScore}, and further propose \textbf{THS}, a learnable automatic evaluator. Our experiments reveal a clear gap between strong performance on conventional generation metrics and true effectiveness under healing-oriented evaluation. Beyond benchmarking, we show that structured score-plus-rationale feedback can directly support generation improvement, and that the HEAL taxonomy provides a useful basis for personalized healing-content delivery. In conclusion, our results establish psychological comfort as a distinct evaluation and optimization target in generative vision, and position HealBench as a strong foundation for future work on affect-aware, human-centered multimodal generation.

\clearpage
\bibliography{sample-base}

\end{document}


\maketitle

\section{Overview}
\label{sec:sup-overview}

This supplementary material provides additional empirical evidence and
technical details that complement the main paper. To foreground the empirical
findings, we first present extended experimental results and analyses, followed
by additional methodological and implementation details, complete experimental
settings, and the full protocol and ethical considerations for the human
evaluation.

The supplementary material is organized as follows:

\begin{itemize}
    \item \textbf{Section one, ``Additional Experimental Results and Analyses,''} presents extended empirical results that support
    and complement the findings reported in the main paper. This section
    includes additional comparisons, robustness analyses, ablation studies,
    qualitative results, and error analyses.

    \item \textbf{Section two, ``Additional Methodological Details,''} provides further explanations of the proposed method, including
    component definitions, design choices, algorithmic procedures, and
    implementation details omitted from the main paper due to space
    constraints.

    \item \textbf{Section three, ``Additional Experimental
    Settings,''} describes the experimental setup in greater detail, including
    datasets, data preprocessing, model and baseline configurations,
    hyperparameters, evaluation metrics, and implementation environments
    required for reproducibility.

    \item \textbf{Section four, ``Human Evaluation Protocol and
    Ethical Considerations,''} presents the complete human-study design,
    including participant recruitment, task procedures, evaluation criteria,
    quality-control mechanisms, statistical analyses, informed consent, and
    relevant ethical considerations.
\end{itemize}

\section{1. Additional Experimental Results and Analyses}
\label{sec:sup-analysis}

This section provides a more detailed analysis of the experimental results reported in the main paper. Rather than introducing additional models, data, or evaluation protocols, we further examine the existing results from the perspectives of evaluation-set representativeness, model-level capability differences, healing-function-specific performance, feedback-based prompt rewriting, automatic scoring, reward-guided optimization, and personalized healing-content delivery. Unless otherwise specified, all models, prompts, evaluation criteria, and experimental settings are identical to those used in the main paper.

\begin{figure}[htbp]
    \centering
    \includegraphics[width=1\linewidth]{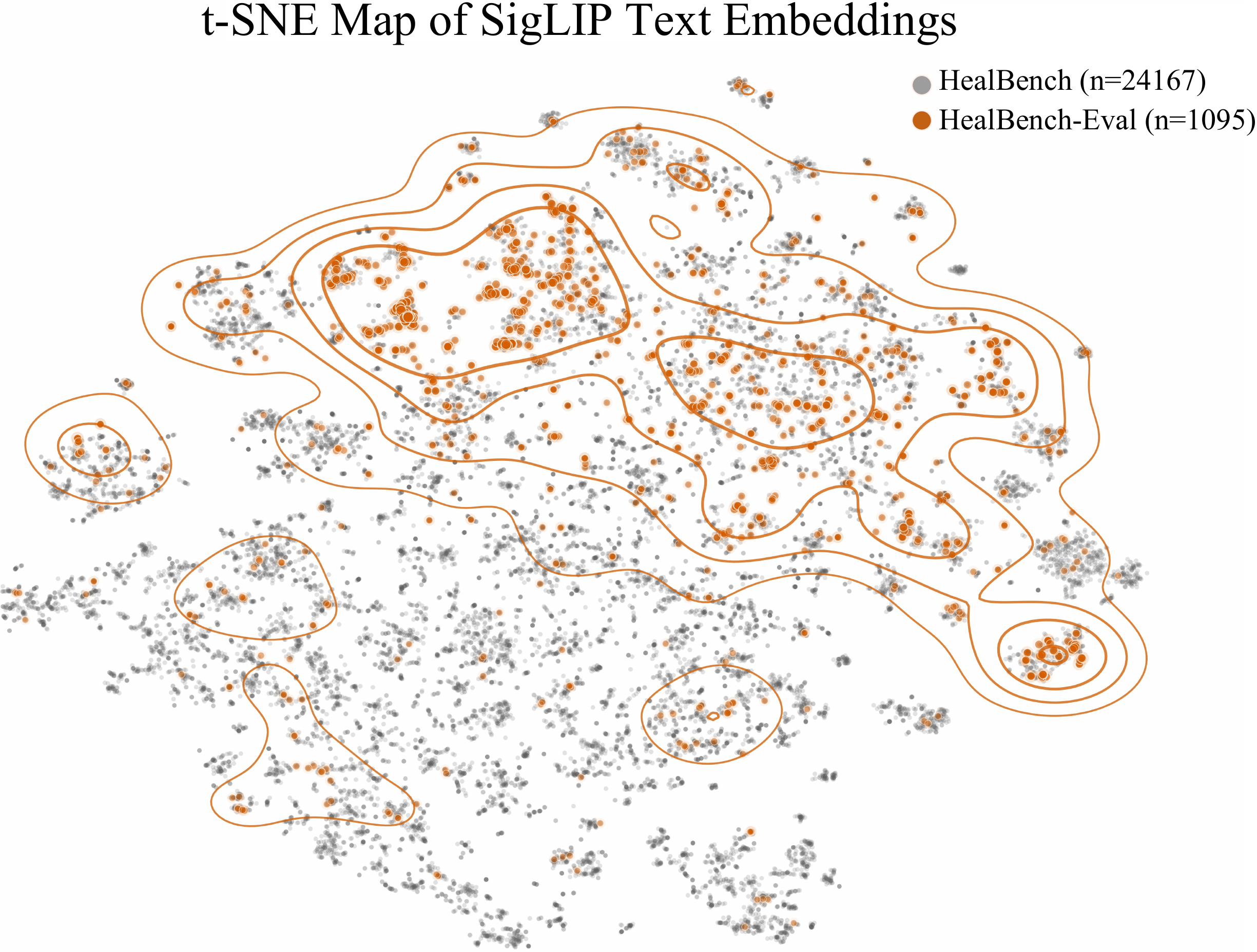}
    \caption{t-SNE visualization of SigLIP embeddings for the full HealBench data and the selected HealBench-Eval. 
}
    \label{fig:tsne}
\end{figure}

\begin{table}[htbp]
\centering
\caption{Quantitative comparison of different text-to-image models across multiple evaluation metrics.}\vspace{-8pt}
\label{tab:t2i_compare_wide}
\scriptsize
\setlength{\tabcolsep}{4pt}
\renewcommand{\arraystretch}{1.05}
\resizebox{\columnwidth}{!}{%
\begin{tabular}{lcccc}
\toprule
Model & LAION Aesthetic $\uparrow$ & CLIPScore $\uparrow$ & PickScore $\uparrow$ & HPSv3 $\uparrow$ \\
\midrule
SD 2.1            & 4.403 & 0.885 & 0.785 & 3.406 \\
SDXL Base 1.0     & 4.631 & 0.885 & 0.790 & 3.602 \\
pixArt-XL    & 4.717 & 0.946 & 0.864 & 12.623 \\
SDXL Turbo     & 4.095 & 0.975 & 0.812 & 5.543 \\
Playground v2.5 & 5.164 & \textbf{0.988} & \textbf{0.899} & 12.712 \\
Hunyuan-Dit         & \textbf{5.271} & 0.976 & 0.872 & 10.991 \\
SD 3 Medium          & 4.841 & 0.965 & 0.863 & 11.515 \\
Kolors          & 4.638 & 0.901 & 0.823 & 9.587 \\
FLUX.1 [dev]      & 4.798 & 0.938 & 0.876 & \textbf{13.585} \\
FLUX.1 [schnell]  & 4.682 & 0.943 & 0.840 & 10.624 \\
SD 3.5 Large         & 4.912 & 0.971 & 0.867 & 11.792 \\
SD 3.5 Medium         & 4.867 & 0.974 & 0.867 & 11.533 \\
sana16          & 4.666 & 0.967 & 0.870 & 12.955 \\
Lumina-Image 2.0  & 4.950 & 0.984 & 0.864 & 11.919 \\
\midrule
GT              & 4.954 & 0.925 & 0.838 & 9.797 \\
\bottomrule
\end{tabular}\vspace{-15pt}
}
\end{table}

\newcommand{\upcell}[2]{%
\makecell[c]{#1\\[-0.25em]\scriptsize\textcolor[rgb]{0.85,0.33,0.10}{$\uparrow$#2}}%
}

\newcommand{\blueupcell}[2]{%
\makecell[c]{#1\\[-0.25em]\scriptsize\textcolor[rgb]{0.00,0.45,0.74}{$\uparrow$#2}}%
}

\begin{table*}[htbp]
\centering
\scriptsize
\setlength{\tabcolsep}{4pt}
\caption{Baseline summary table. All metrics are positive indicators. For prompt-rewrite models, the orange mark denotes the improvement over the corresponding non-rewrite base model. For image editing models, the blue mark denotes the improvement over the corresponding original model.}\vspace{-8pt}
\label{tab:baseline_summary}
\resizebox{\textwidth}{!}{%
\begin{tabular}{lcccccccccc}
\toprule
Model & Overall & Text-Image Alignment & Heal Accuracy & Image Quality & Natural Restoration & Atmospheric Immersion & Shelter and Safety & Living Companionship & Rhythmic Focus & Awe and Transcendence \\
\midrule

\multicolumn{11}{c}{\textit{Diffusion Models}} \\
\midrule
SD 2.1 & 19.390 & 5.172 & 4.820 & 4.578 & 18.754 & 19.678 & 18.484 & 21.780 & 19.412 & 20.147 \\
SDXL Base 1.0 & 20.723 & 5.319 & 5.173 & 5.057 & 20.258 & 21.269 & 21.915 & 24.284 & 19.859 & 20.481 \\
PixArt-XL & 25.080 & 6.435 & 6.169 & 6.308 & 24.953 & 24.832 & 25.115 & 27.787 & 25.036 & 24.844 \\
SDXL Turbo & 18.765 & 5.419 & 4.517 & 4.312 & 17.372 & 18.689 & 17.536 & 19.837 & 20.549 & 18.695 \\
Playground v2.5 & 25.088 & 6.479 & 6.094 & 6.422 & 24.551 & 25.304 & 24.449 & 28.177 & 25.162 & 25.253 \\
Hunyuan-DiT & 25.020 & 6.378 & 6.178 & 6.286 & 24.341 & 24.849 & 24.286 & 28.057 & 25.427 & 25.347 \\
Kolors & 23.535 & 6.005 & 5.866 & 5.797 & 23.586 & 23.924 & 22.832 & 26.369 & 23.267 & 23.255 \\
Sana-1.6B & 24.419 & 6.386 & 5.917 & 6.201 & 23.748 & 25.044 & 24.291 & 29.624 & 23.821 & 24.541 \\

\midrule
\multicolumn{11}{c}{\textit{Flow Matching Models}} \\
\midrule
SD 3 Medium & 23.465 & 6.121 & 5.731 & 5.882 & 22.556 & 23.691 & 21.617 & 28.560 & 23.462 & 24.780 \\
FLUX.1 [dev] & 24.836 & 6.289 & 6.152 & 6.243 & 24.211 & 24.195 & 23.489 & 29.759 & 25.424 & 25.222 \\
FLUX.1 [schnell] & 22.717 & 5.954 & 5.561 & 5.642 & 22.080 & 21.632 & 21.108 & 27.830 & 23.225 & 23.913 \\
SD 3.5 Large & 24.571 & 6.346 & 6.008 & 6.210 & 23.694 & 24.381 & 23.160 & 29.057 & 25.106 & 25.154 \\
SD 3.5 Medium & 24.164 & 6.321 & 5.884 & 6.075 & 23.285 & 24.131 & 22.391 & 28.730 & 24.266 & 25.620 \\
Lumina-Image 2.0 & 23.169 & 6.132 & 5.627 & 5.782 & 22.483 & 22.855 & 22.035 & 28.447 & 23.433 & 23.602 \\

\midrule
\multicolumn{11}{c}{\textit{Prompt-Rewrite Models}} \\
\midrule
Hunyuan-DiT (Rewrite)
& \upcell{26.631}{1.612}
& \upcell{6.617}{0.239}
& \upcell{6.711}{0.533}
& \upcell{6.592}{0.306}
& \upcell{26.841}{2.500}
& \upcell{25.843}{0.994}
& \upcell{26.336}{2.050}
& \upcell{29.546}{1.489}
& \upcell{27.183}{1.757}
& \upcell{25.574}{0.226} \\
FLUX.1 [schnell] (Rewrite)
& \upcell{24.514}{1.797}
& \upcell{6.328}{0.374}
& \upcell{6.051}{0.490}
& \upcell{6.084}{0.442}
& \upcell{24.411}{2.331}
& \upcell{23.090}{1.459}
& \upcell{23.942}{2.835}
& \upcell{28.794}{0.965}
& \upcell{25.043}{1.818}
& \upcell{24.552}{0.638} \\
SD 3.5 Large (Rewrite)
& \upcell{26.149}{1.578}
& \upcell{6.638}{0.293}
& \upcell{6.471}{0.464}
& \upcell{6.568}{0.358}
& \upcell{25.851}{2.157}
& \upcell{24.993}{0.611}
& \upcell{25.825}{2.664}
& \upcell{31.057}{2.000}
& \upcell{26.680}{1.575}
& \upcell{25.836}{0.683} \\
SD 3.5 Medium (Rewrite)
& \upcell{25.598}{1.434}
& \upcell{6.556}{0.235}
& \upcell{6.322}{0.438}
& \upcell{6.398}{0.323}
& \upcell{25.071}{1.786}
& \upcell{24.683}{0.552}
& \upcell{24.554}{2.163}
& \upcell{29.979}{1.248}
& \upcell{26.179}{1.912}
& \upcell{26.046}{0.426} \\

\midrule
\multicolumn{11}{c}{\textit{Image Editing Models}} \\
\midrule
Pix2Pix (Original) & 20.666 & 5.454 & 5.081 & 5.050 & 20.349 & 21.401 & 20.832 & 23.837 & 20.832 & 19.059 \\
Pix2Pix (Rewrite)
& \blueupcell{22.802}{2.136}
& \blueupcell{5.853}{0.399}
& \blueupcell{5.664}{0.583}
& \blueupcell{5.620}{0.570}
& \blueupcell{23.047}{2.698}
& \blueupcell{23.193}{1.792}
& \blueupcell{23.649}{2.817}
& \blueupcell{25.674}{1.837}
& \blueupcell{22.719}{1.887}
& \blueupcell{20.606}{1.547} \\
SDXL-Pix2Pix (Original) & 19.307 & 5.137 & 4.722 & 4.727 & 18.612 & 19.576 & 18.351 & 23.780 & 19.272 & 19.733 \\
SDXL-Pix2Pix (Rewrite)
& \blueupcell{22.053}{2.746}
& \blueupcell{5.711}{0.574}
& \blueupcell{5.445}{0.723}
& \blueupcell{5.453}{0.726}
& \blueupcell{22.111}{3.499}
& \blueupcell{22.575}{2.999}
& \blueupcell{21.464}{3.113}
& \blueupcell{26.255}{2.475}
& \blueupcell{21.908}{2.636}
& \blueupcell{20.917}{1.184} \\

\midrule
\multicolumn{11}{c}{\textit{Ground Truth}} \\
\midrule
Real Images & 24.528 & 6.375 & 5.995 & 6.163 & 23.186 & 24.119 & 25.205 & 27.901 & 25.216 & 24.503 \\
\bottomrule
\end{tabular}%
}
\vspace{-10pt}
\end{table*}

\subsection{1.1 Additional Analysis of HealBench-Eval}
\label{sec:sup-eval-analysis}

HealBench-Eval is designed as a compact evaluation set that preserves the categorical and visual diversity of the larger HealBench data pool while substantially reducing the cost of model comparison. Instead of directly applying random sampling over the complete collection, we perform category-conditioned representative selection within each of the 92 valid category--subcategory pairs. This design prevents categories with larger source pools from dominating the evaluation set and ensures that fine-grained healing functions remain represented during benchmarking.

Figure~\ref{fig:tsne} visualizes the SigLIP embedding distributions of the larger data pool and the selected HealBench-Eval instances. The selected samples are not concentrated exclusively in a small number of dense embedding regions. Instead, they remain distributed across the major semantic regions and several peripheral regions of the original representation space. This pattern suggests that the construction process retains both common visual patterns and less frequent category-specific variations, rather than selecting only prototypical or visually homogeneous examples.

The clustering-based construction is particularly important for HealBench because images belonging to the same nominal category may express substantially different visual and affective cues. For example, two images within \emph{Shelter and Safety} may both depict indoor environments while differing in lighting, spatial openness, perceived warmth, visual clutter, and the degree of psychological security they convey. Similarly, samples within \emph{Natural Restoration} may vary from quiet forest paths to expansive coastal landscapes, which can produce different levels of calmness, cognitive demand, and emotional distance. Representative selection in the embedding space therefore provides a more informative evaluation set than selecting samples solely according to category labels or aggregate HealScores.

The visualization in Figure~\ref{fig:tsne} should be interpreted as qualitative evidence of representational coverage rather than as proof that the two distributions are identical. Nevertheless, together with the category-conditioned sampling procedure, it supports the use of HealBench-Eval as a controlled and computationally manageable test set for comparing multiple generation models, prompt-rewriting strategies, image-editing systems, and automatic evaluators under a shared protocol.

\subsection{1.2 Fine-Grained Analysis of Baseline Models}
\label{sec:sup-baseline-analysis}

Tables~\ref{tab:t2i_compare_wide} and~\ref{tab:baseline_summary} evaluate the same generation systems from two complementary perspectives. Table~\ref{tab:t2i_compare_wide} reports conventional metrics related to aesthetics, text--image correspondence, and generic human preference, whereas Table~\ref{tab:baseline_summary} reports the healing-oriented evaluation defined by HealBench. Comparing the two tables reveals that strong performance under conventional evaluation does not uniquely determine strong performance under perceived psychological comfort.

For example, FLUX.1 [dev] achieves the highest HPSv3 score among the evaluated systems in Table~\ref{tab:t2i_compare_wide}, but its Overall HealScore in Table~\ref{tab:baseline_summary} remains below those of Playground v2.5, PixArt-XL, and Hunyuan-DiT. Hunyuan-DiT obtains the highest LAION Aesthetic score and also performs strongly under HealBench, whereas Playground v2.5 achieves the highest CLIPScore and PickScore and narrowly obtains the best Overall HealScore among the original generation baselines. These partially overlapping but non-identical rankings indicate that semantic fidelity, aesthetic appeal, generic preference, and psychological comfort capture related yet distinct properties of generated images.

Among the original baselines, Playground v2.5, PixArt-XL, and Hunyuan-DiT achieve Overall HealScores of 25.088, 25.080, and 25.020, respectively. Their overall results are extremely close, but their dimension-level profiles are not identical. Playground v2.5 obtains 6.479 for Text-Image Alignment, 6.094 for Heal Accuracy, and 6.422 for Image Quality, while PixArt-XL obtains 6.435, 6.169, and 6.308 on the same dimensions. Hunyuan-DiT achieves a slightly more balanced profile of 6.378, 6.178, and 6.286. These results illustrate why a single aggregate ranking is insufficient for diagnostic analysis: models with nearly identical overall scores may arrive at those scores through different combinations of semantic accuracy, healing relevance, and visual quality.

A consistent pattern among many of the stronger models is that Heal Accuracy remains lower than Text-Image Alignment and Image Quality. For example, SD 3.5 Medium achieves 6.321 in Text-Image Alignment and 6.075 in Image Quality, but only 5.884 in Heal Accuracy. Similarly, Playground v2.5 reaches 6.479 and 6.422 in alignment and quality, respectively, while obtaining 6.094 in Heal Accuracy. These gaps suggest that current models are often capable of producing images that are visually polished and semantically appropriate without fully expressing the restorative, supportive, or psychologically comforting cues associated with the target healing function.

The model-family comparison also indicates that no single generation paradigm directly guarantees stronger healing-oriented performance. Several diffusion-based systems, including Playground v2.5, PixArt-XL, and Hunyuan-DiT, outperform the evaluated flow-matching systems in Overall HealScore, while FLUX.1 [dev] remains highly competitive and performs particularly well for selected healing functions. Faster variants also tend to underperform their corresponding higher-capacity counterparts in the current evaluation. FLUX.1 [schnell], for instance, obtains an Overall HealScore of 22.717 compared with 24.836 for FLUX.1 [dev], while SDXL Turbo obtains 18.765 compared with 20.723 for SDXL Base 1.0. These descriptive results suggest that generation speed and generic visual capability do not necessarily preserve the affective details required for psychological comfort.

The real-image results in Table~\ref{tab:baseline_summary} provide a useful reference rather than a strict upper bound. Some generated models exceed the mean real-image score in selected dimensions or healing classes, while remaining weaker in others. This comparison should not be interpreted as evidence that generated images are categorically more psychologically comforting than real images, because the two groups differ in source, sampling, and prompt conditioning. Instead, the reference results show that advanced generators can approach the average quality of curated real imagery under specific healing-oriented criteria, while still exhibiting uneven capability across dimensions and categories.

\subsection{1.3 Performance Across Healing Functions}
\label{sec:sup-class-analysis}

The six HEAL classes in Table~\ref{tab:baseline_summary} reveal substantial function-specific variation that is not visible from the Overall HealScore alone. No original model achieves the highest score across all six classes, indicating that healing-oriented generation is not a single homogeneous capability. Instead, different restorative functions appear to rely on distinct visual structures, semantic cues, and affective properties.

For \emph{Natural Restoration}, PixArt-XL achieves the strongest result among the original models with a score of 24.953, followed by Playground v2.5 and Hunyuan-DiT. This category may appear relatively close to conventional landscape generation, but generating an aesthetically attractive natural scene is not sufficient by itself. The image must also avoid excessive visual intensity, dramatic tension, or compositional complexity that may weaken its restorative character. The results therefore reflect not only a model's ability to generate nature, but also its ability to present nature in a calm, accessible, and psychologically supportive manner.

For \emph{Atmospheric Immersion}, Playground v2.5 obtains the highest original-model score of 25.304, while Sana-1.6B and Hunyuan-DiT also perform strongly. This class depends heavily on ambient properties such as light, weather, color transitions, spatial depth, and sensory texture. A model may correctly depict rain, snow, candlelight, or a nighttime street while still producing an atmosphere that feels cold, threatening, visually harsh, or overly cinematic. HealBench distinguishes these affective differences from basic category recognition.

For \emph{Shelter and Safety}, PixArt-XL achieves 25.115, the strongest result among the original models, while the remaining systems show comparatively larger variation. This class requires the simultaneous expression of physical enclosure, emotional warmth, visual stability, and the absence of threatening or oppressive cues. Indoor spaces that are technically correct but excessively dark, empty, crowded, or geometrically distorted may fail to communicate psychological security. The relatively uneven performance across models suggests that shelter-related comfort requires a more complex combination of spatial and affective reasoning than the presence of a recognizable indoor scene alone.

\emph{Living Companionship} produces some of the highest class-level scores in the benchmark. FLUX.1 [dev] reaches 29.759, Sana-1.6B reaches 29.624, and SD 3.5 Large reaches 29.057. Living beings provide direct and recognizable affiliative cues, including warmth, vitality, attachment, and social presence, which may be easier for current models to express than more abstract restorative functions. However, the category is not automatically successful whenever an animal or person is present. Unnatural anatomy, inappropriate gaze, awkward interaction, crowding, or ambiguous emotional expression can still reduce perceived comfort.

For \emph{Rhythmic Focus}, Hunyuan-DiT and FLUX.1 [dev] obtain nearly identical leading scores of 25.427 and 25.424. This class emphasizes low-demand, process-oriented attention, including activities such as studying, coding, coffee preparation, baking, and fluid or particle-based visual processes. Successful images must communicate sustained but non-stressful engagement. Excessive clutter, strong task pressure, dramatic motion, or visually confusing object arrangements can transform a potentially calming activity into a cognitively demanding scene.

For \emph{Awe and Transcendence}, SD 3.5 Medium achieves the highest original-model result of 25.620. This category illustrates an important distinction between visual impressiveness and psychological comfort. Large-scale landscapes, galaxies, auroras, and surreal spaces can evoke expansion and transcendence, but they may also create distance, threat, loneliness, or overwhelming visual intensity. The strongest systems must therefore balance grandeur with visual coherence and emotional accessibility.

Overall, the class-wise results demonstrate that psychological comfort is strongly category dependent. Living Companionship benefits from explicit affiliative subjects, while Shelter and Safety requires more subtle integration of spatial structure, lighting, and psychological security. Awe and Transcendence can benefit from the visual strengths of modern generators but remains vulnerable to excessive intensity. These differences support the need for a fine-grained benchmark that reports performance across multiple healing functions rather than reducing all restorative imagery to a single undifferentiated preference score.

\begin{figure}[htbp]
    \centering
    \includegraphics[width=1.0\linewidth]{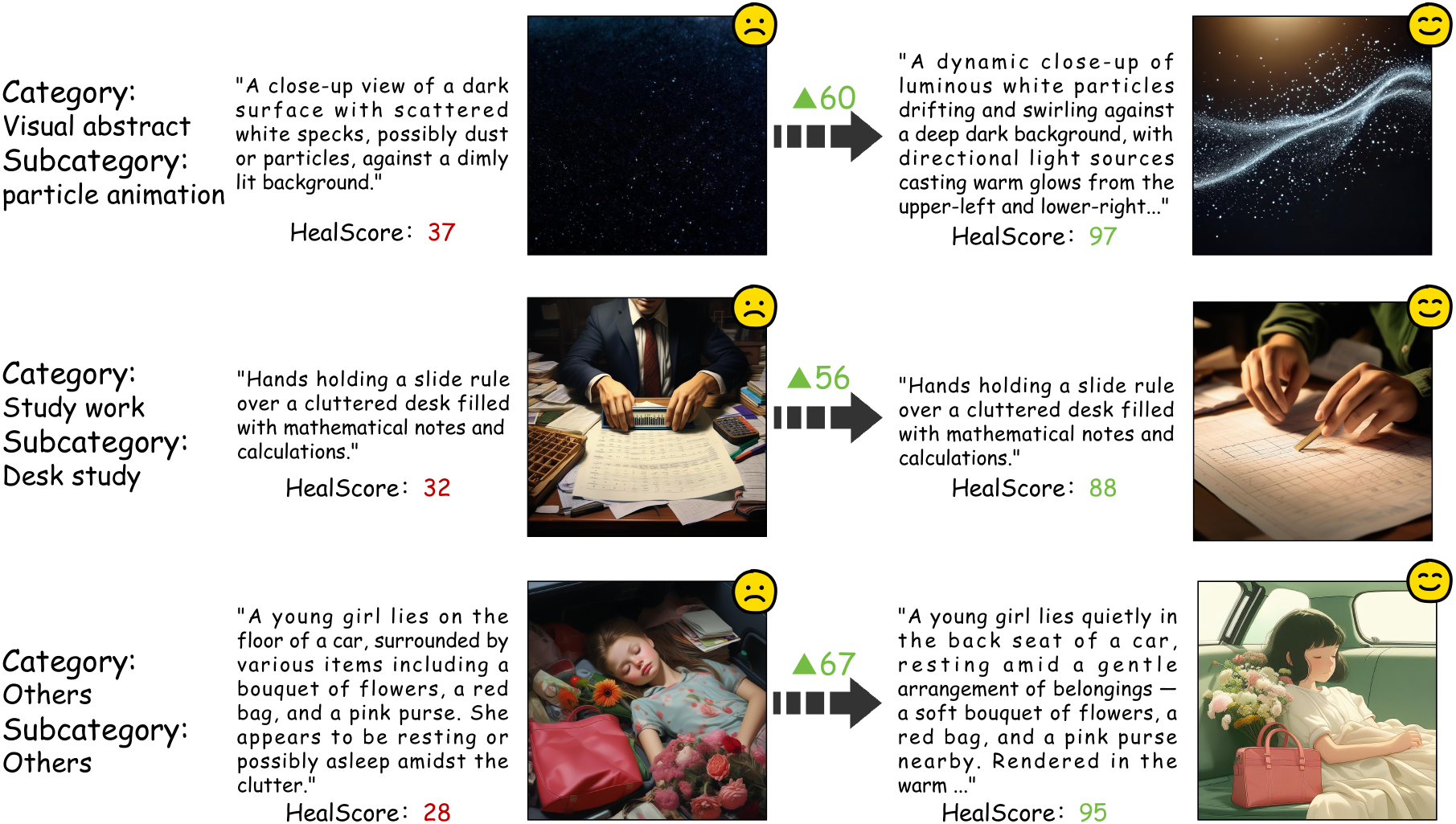} \vspace{-8pt}
    \caption{Qualitative examples of our prompt rewriting strategy. Using the same image generation model, targeted prompt rewriting markedly improves generation outcomes, yielding more coherent and safer images with substantially higher HealScores across multiple categories.} 
    \vspace{-10pt}
    \label{fig:re_image}      
\end{figure}

\subsection{1.4 Additional Analysis of Prompt Rewriting}
\label{sec:sup-rewrite-analysis}

The prompt-rewriting results in Table~\ref{tab:baseline_summary} show that structured score-plus-rationale feedback consistently improves all evaluated text-to-image and image-editing systems. Among the four text-to-image models, the Overall HealScore increases by 1.612 for Hunyuan-DiT, 1.797 for FLUX.1 [schnell], 1.578 for SD 3.5 Large, and 1.434 for SD 3.5 Medium. These gains are observed despite substantial differences in the architectures and original performance levels of the models, suggesting that the feedback is not restricted to a single generator family.

For all four text-to-image models, the largest dimension-level improvement occurs in Heal Accuracy. Hunyuan-DiT improves by 0.533 in Heal Accuracy, compared with gains of 0.239 in Text-Image Alignment and 0.306 in Image Quality. FLUX.1 [schnell] improves by 0.490 in Heal Accuracy, while SD 3.5 Large and SD 3.5 Medium improve by 0.464 and 0.438, respectively. This pattern is consistent with the intended role of the feedback mechanism: the rewriting process does not merely add visual details or improve linguistic specificity, but places greater emphasis on category-relevant restorative and emotionally supportive cues.

The class-level improvements are also uneven, indicating that prompt rewriting is particularly valuable for healing functions whose affective requirements are difficult to specify in short captions. Natural Restoration and Shelter and Safety show several of the largest gains. For example, rewriting improves Hunyuan-DiT by 2.500 on Natural Restoration and 2.050 on Shelter and Safety. FLUX.1 [schnell] improves by 2.331 and 2.835 on the same classes, while SD 3.5 Large improves by 2.157 and 2.664. These results suggest that explicit feedback can help prompts move beyond generic category descriptions by specifying softer lighting, reduced visual tension, clearer spatial safety, natural compositional balance, and other cues that contribute to perceived comfort.

Prompt rewriting also benefits instruction-guided image editing. Pix2Pix improves by 2.136 in Overall HealScore, while SDXL-Pix2Pix improves by 2.746. Both editing systems obtain positive changes across all three evaluation dimensions and all six HEAL classes. The gains are especially pronounced for Natural Restoration, Atmospheric Immersion, and Shelter and Safety, where the rewriting process can provide more explicit guidance about atmosphere, lighting, clutter reduction, and emotionally supportive visual structure. The consistency across text-to-image generation and image editing suggests that the score-plus-rationale paradigm can function as a task-level feedback mechanism rather than as a model-specific prompt heuristic.

The qualitative examples in Figure~\ref{fig:re_image} further illustrate how rewritten prompts affect the generated content. Across the presented examples, the rewritten prompts produce more coherent subject arrangements, reduce ambiguous or visually tense elements, and strengthen cues associated with calmness, safety, and restorative attention. The corresponding HealScore improvements of 67, 56, and 60 points show that the observed changes are not limited to minor visual refinements. Instead, the rewritten prompts can substantially alter whether the resulting image expresses the intended healing semantics.

Because the current experiment does not include a generic prompt-expansion baseline or separate score-only and rationale-only conditions, the results should be interpreted as evidence for the utility of the complete score-plus-rationale rewriting pipeline rather than as an isolated causal estimate of the rationale component. Nevertheless, the consistent improvements across models, evaluation dimensions, healing functions, and generation tasks demonstrate that structured diagnostic feedback can be translated into actionable modifications of visual generation prompts.

\begin{table}[htbp]
\centering 
\scriptsize
\setlength{\tabcolsep}{2.5pt}
\renewcommand{\arraystretch}{0.95}
\caption{THS vs. generic scorers on pairwise healing-preference ranking.}\vspace{-8pt}
\resizebox{\columnwidth}{!}{%
\begin{tabular}{lccc}
\toprule 
Model & Acc. $\uparrow$ & Kendall's $\tau$ $\uparrow$ & Spearman's $\rho$ $\uparrow$ \\
\midrule
PickScore      & 0.5593          & 0.1525          & 0.2254          \\
LAION Aesthetic      & 0.5507          & 0.1077          & 0.1595          \\
CLIPScore      & 0.5413          & 0.1272          & 0.1882          \\
HPSv3          & 0.5013          & 0.0379          & 0.0595          \\
THS  & \textbf{0.8527} & \textbf{0.7741} & \textbf{0.9223} \\
\bottomrule
\end{tabular}
}
\label{tab:rm_compare}
\vspace{-8pt}
\end{table}

\subsection{1.5 Additional Analysis of THS}
\label{sec:sup-ths-analysis}

Table~\ref{tab:rm_compare} compares THS with four generic image-scoring models under the same pairwise healing-preference protocol. THS achieves an accuracy of 0.8527, a Kendall's $\tau$ of 0.7741, and a Spearman's $\rho$ of 0.9223. In contrast, the strongest generic baseline reaches only 0.5593 in accuracy, 0.1525 in Kendall's $\tau$, and 0.2254 in Spearman's $\rho$. The large margin across all three metrics indicates that the advantage of THS is not limited to isolated pairwise decisions, but extends to the broader consistency of the induced ranking.

The three reported metrics capture complementary aspects of scorer behavior. Pairwise accuracy measures whether a scorer selects the same preferred image as the reference judgment for an individual pair. Kendall's $\tau$ evaluates the consistency of relative ordering across comparisons, while Spearman's $\rho$ measures agreement between the overall ranking trends. The simultaneously high values obtained by THS therefore suggest that it provides both reliable local decisions and a stable global ordering under the current evaluation protocol.

The weaker performance of generic scorers is consistent with their original design objectives. CLIPScore primarily measures semantic correspondence between an image and a textual description, while LAION Aesthetic emphasizes visual appeal. PickScore and HPSv3 model broad human preference, which may reward realism, detail, composition, or prompt fidelity without explicitly distinguishing whether an image conveys psychological safety, companionship, restoration, or low-demand attention. These properties are relevant to overall visual quality, but they do not fully represent the target construct evaluated by HealBench.

THS differs from these scorers by learning dimension-specific supervision for Text-Image Alignment, Heal Accuracy, and Image Quality. This structure allows it to preserve the supporting constraints of semantic correctness and visual quality while explicitly modeling the target healing dimension. Consequently, THS serves not only as an alternative aggregate scorer, but also as a scalable approximation of HealBench's structured evaluation protocol.

The results in Table~\ref{tab:rm_compare} establish that THS is better aligned with healing-oriented preference than the evaluated generic scorers under the current benchmark and comparison setting. They should not be interpreted as evidence that THS universally generalizes to every image domain, psychological population, or unseen healing concept. Within the scope of HealBench, however, THS provides a substantially more suitable automatic signal for large-scale model comparison, result filtering, and downstream optimization.

\begin{table}[htbp]
\centering
\caption{Main results of the base SD 3.5 Medium model and its Flow-GRPO fine-tuned variant. Higher values are highlighted in blue.}\vspace{-8pt}
\label{tab:sd35_flowgrpo_main}
\setlength{\tabcolsep}{3.2pt}
\renewcommand{\arraystretch}{1.15}
\small
\begin{tabular}{lcccc}
\toprule
Model & Overall & Semantic & Heal & Quality \\
\midrule
SD 3.5 Medi (base)
& 24.16
& 6.32
& 5.88
& 6.08 \\
SD 3.5 Medi + Flow-GRPO
& \textbf{25.67}
& \textbf{6.48}
& \textbf{6.45}
& \textbf{6.28} \\
\bottomrule
\end{tabular}
\vspace{-10pt}
\end{table}

\subsection{1.6 Exploratory Analysis of THS-Guided Optimization}
\label{sec:sup-rl-analysis}

The experiment in Table~\ref{tab:sd35_flowgrpo_main} examines whether THS can function as an optimization signal in addition to serving as an evaluator. Starting from SD 3.5 Medium, the Flow-GRPO fine-tuned model improves the Overall HealScore from 24.16 to 25.67. The dimension-level scores increase from 6.32 to 6.48 for Semantic Alignment, from 5.88 to 6.45 for Heal Accuracy, and from 6.08 to 6.28 for Image Quality.

The largest component-level improvement occurs in Heal Accuracy, which increases by 0.57. This result is consistent with the intended role of THS as a healing-oriented reward signal. At the same time, Semantic Alignment and Image Quality also improve by 0.16 and 0.20, respectively. The optimization therefore does not appear to increase the target healing dimension by directly sacrificing the two supporting dimensions in the reported evaluation.

These results extend the role of HealBench from passive benchmarking to optimization-oriented feedback. A benchmark is more actionable when its evaluation signal can identify failure modes and support improvement of the underlying generator. The Flow-GRPO experiment provides initial evidence that the structured signal learned by THS can influence the generation model toward outputs that receive stronger evaluations across all three HealBench dimensions.

This experiment remains exploratory. It evaluates a single base model, one fine-tuning method, and one reward formulation, and therefore does not establish that THS-guided optimization will transfer uniformly across architectures or training algorithms. It also does not independently rule out all forms of reward over-optimization. The result should instead be understood as a feasibility demonstration showing that a healing-oriented evaluator can be incorporated into a direct model-optimization pipeline and produce measurable improvements on the target evaluation dimensions.

\begin{table}[htbp]
\centering
\small
\setlength{\tabcolsep}{4pt}
\renewcommand{\arraystretch}{1.1}
\caption{User study on personalized healing delivery with 24 subthreshold-depression college students. 
Participants in the personalized condition could choose among the six HEAL classes, while those in the random condition viewed randomly assigned healing images. 
Mood change was measured with HAMD-17 after one hour of viewing. }\vspace{-8pt}
\label{tab:user_study_delivery}
\resizebox{\columnwidth}{!}{%
\begin{tabular}{lccc}
\toprule
Condition & $N$ & Improved Ratio (\%) $\uparrow$ & HAMD-17 $\downarrow$ \\
\midrule
Random playback & 12 & 41.7 & -1.27 \\
Personalized class selection & 12 & 75.0 & -3.42 \\
\midrule
Difference / Gain & -- & +29.5 & -2.15 \\
\bottomrule
\end{tabular}
}
\end{table}

\subsection{1.7 Additional Interpretation of the User Study}
\label{sec:sup-user-study-analysis}

Table~\ref{tab:user_study_delivery} reports a preliminary user study involving 24 college students with subthreshold depressive symptoms. Participants were assigned either to a random-playback condition or to a personalized condition in which they selected one of the six HEAL classes. The personalized condition therefore tests whether the functional taxonomy developed for benchmark construction can also support preference-sensitive delivery of psychologically comforting visual content.

The personalized class-selection condition achieves a reported improvement ratio of 75.0\%, compared with 41.7\% for random playback. It also produces a larger reduction in HAMD-17 score, with an average change of $-3.42$ compared with $-1.27$ in the random condition. The resulting differences of 29.5 percentage points in improvement ratio and 2.15 points in score reduction both favor personalized selection.

These results provide preliminary support for the practical relevance of the six HEAL classes. The taxonomy is not used solely to organize benchmark samples or report model-level performance. It also offers participants a meaningful interface for selecting the type of visual experience they currently prefer, such as natural restoration, atmospheric immersion, shelter and safety, living companionship, rhythmic focus, or awe and transcendence. The improved outcome in the personalized condition suggests that different users may benefit from different restorative functions, and that a single randomly assigned category may not adequately reflect individual affective needs.

The user study should nevertheless be interpreted as an exploratory feasibility study rather than as evidence of clinical efficacy. The sample size is small, the participant population is restricted to college students with subthreshold symptoms, and the observation period is short. The results do not establish long-term therapeutic effects, clinical treatment benefits, or generalization to broader populations. Their primary contribution is to provide initial evidence that the HEAL taxonomy may support personalized delivery and that healing-oriented image evaluation has potential relevance beyond offline benchmark comparison.

\subsection{1.8 Summary of Additional Findings}
\label{sec:sup-analysis-summary}

The additional analyses reinforce five central observations from the main paper. First, conventional semantic, aesthetic, and preference metrics overlap with but do not fully determine healing-oriented model performance. Second, the tri-dimensional results reveal that Heal Accuracy often remains a bottleneck even when Text-Image Alignment and Image Quality are relatively strong. Third, model capability varies substantially across the six healing functions, and no original model dominates all categories. Fourth, score-plus-rationale prompt rewriting consistently improves the evaluated text-to-image and image-editing systems, with particularly clear gains in Heal Accuracy and in categories that depend on subtle atmospheric or safety-related cues. Finally, THS provides a substantially stronger healing-oriented ranking signal than generic scorers and shows preliminary potential as a reward for direct generator optimization. Together with the exploratory user study, these findings support HealBench as a diagnostic, actionable, and human-centered framework for evaluating perceived psychological comfort in visual generation.

\section{Additional Methodological Details}
\label{sec:sup-method}
\subsection{Data Collection and Annotation}
\begin{figure*}[t]
    \centering
    \includegraphics[width=1\textwidth]{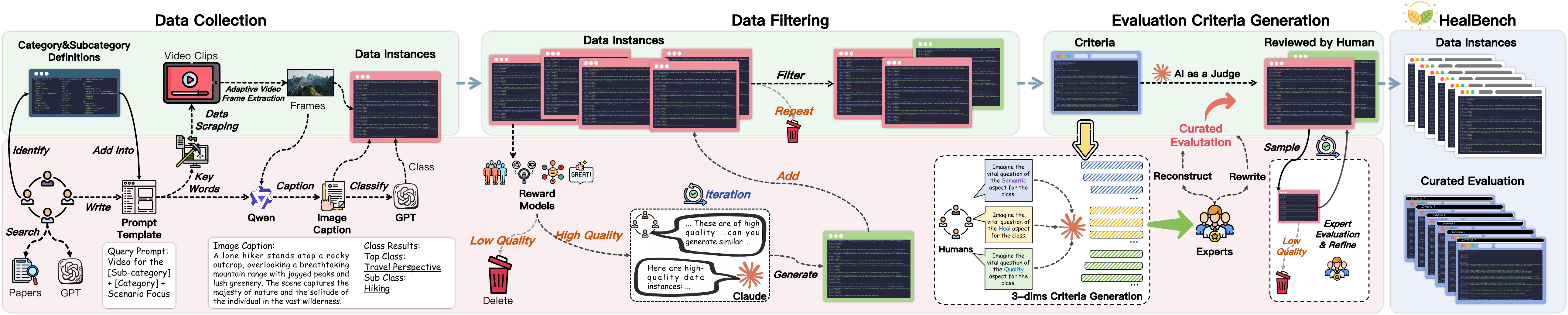}
    \caption{Overview of the HealBench construction pipeline. The benchmark is built through three stages: data collection, reward-model-based filtering and augmentation, and tri-dimensional evaluation-criteria generation. Human and expert review is incorporated throughout the pipeline to ensure the quality of both data instances and evaluation criteria.}
    \label{fig:healbench}
    \vspace{-8pt}
\end{figure*}

We first present the construction process of the dataset. The data collection and processing pipeline of HealBench is illustrated in Fig.~\ref{fig:healbench}. We invited psychology experts and constructed a total of 19 coarse-grained categories and 102 fine-grained sub-categories based on multiple human visual preferences. We denote a category pair as $\{T_i,S_{i,j}\}$, where $T_i$ represents a coarse-grained category, $i$ belongs to $0$--$18$, and $S_{i,j}$ represents a fine-grained sub-category, where $j$ belongs to $0$--the number of fine-grained sub-categories under the coarse-grained category $T_i$. Next, we use the following strings and their augmented variants as keywords to crawl relevant videos from YouTube, Bilibili, and other websites:
\begin{figure*}[!t]
    \centering
    \includegraphics[width=\textwidth]{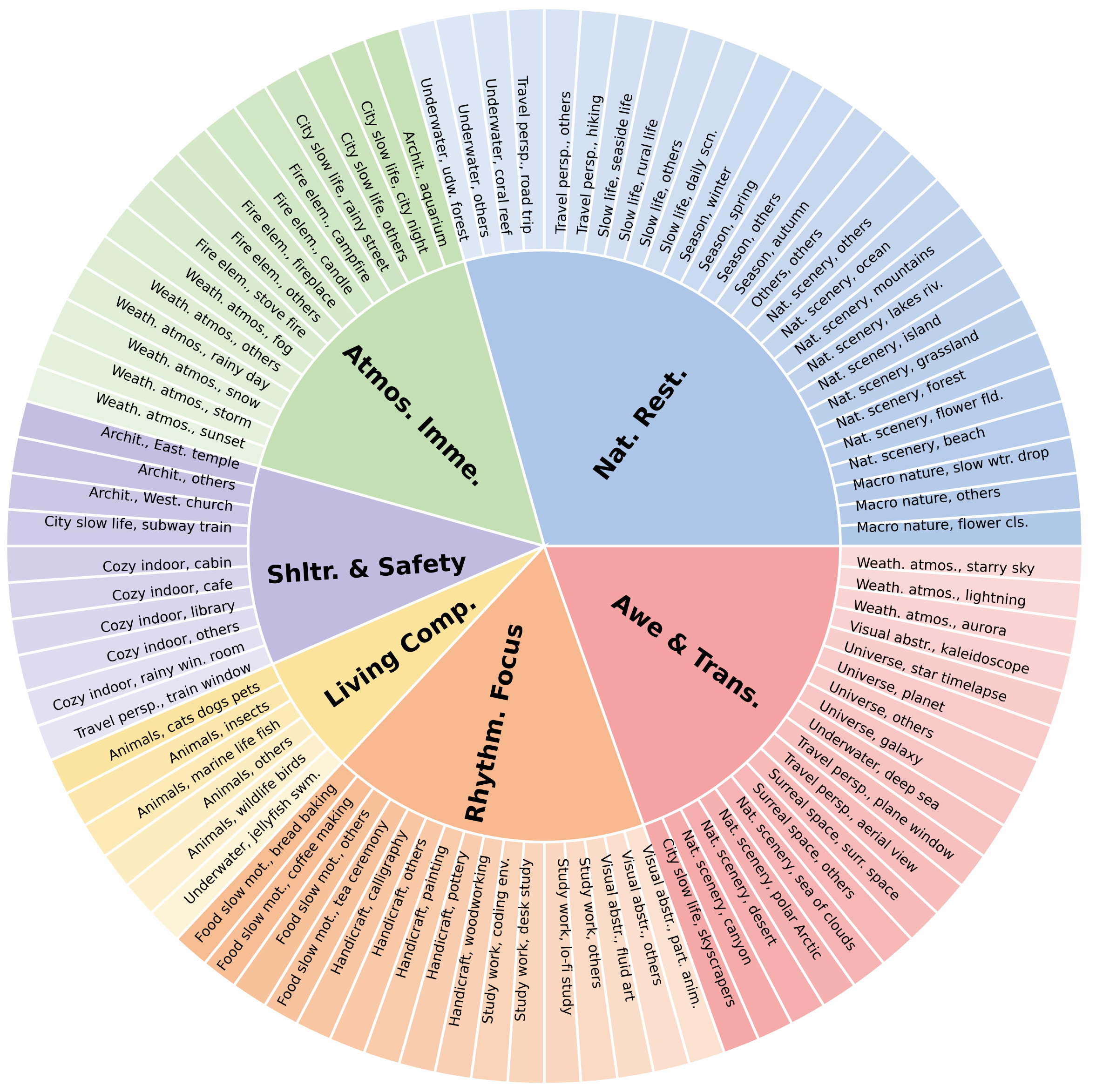}
    \caption{Distribution of the 92 valid subcategory pairs in HEALBENCH, grouped into six major classes according to their psychological soothing effects.}
    \label{fig:pie}
\end{figure*}
\begin{itemize}
    \item relaxing/therapeutic/soothing $\{T_i\}$ $\{S_{i,j}\}$ video
    \item $\{T_i\}$ $\{S_{i,j}\}$ for meditation/sleep/stress relief
    \item ....
\end{itemize}
We collected a total of 358 videos. For each video, we adopt a heuristic frame extraction method: if the video duration is shorter than 10 minutes, the camera transitions in the video may be relatively frequent, and we extract one frame every 2 seconds. If the video duration is longer than 10 minutes, the video may contain a large amount of time repeating the same or similar scenes, and we extract one frame every 10 seconds. We obtain a raw image set ($D_{\mathrm{raw}}$) of approximately 100K frames in total. Next, we use the multimodal vision model Qwen2.5-VL-7B-Instruct to generate a caption for each image. We design the prompt as:

\begin{promptbox}
\small
Write a caption for this image, you should describe the content of the picture to facilitate the subsequent generation of an image using this text as a prompt. Do not generate anything other than the caption.
\end{promptbox}

It is worth noting that the category pair of an image cannot be simply determined according to the keywords used during retrieval. For example, in a video named ``healing moments in nature'', there may be segments related to lakes, forests, and grasslands, and the images obtained from these segments cannot simply be labeled as ``nature'', which would undoubtedly lead to information loss. Due to the large data scale, for image category annotation, we use Claude-Sonnet 4.6 + manual spot-checking to complete the process. The input is designed as:

\begin{promptbox}
You are a strict image scene classifier.

Classify the image into one top-level category and one sub-category.

Taxonomy:
\{category\_block\}

Rules:\\
1) top must be exactly one top category from taxonomy.\\
2) sub must be exactly one sub category under that top.\\
3) If uncertain, choose the closest category.\\
4) If none is suitable, choose top=Others and sub=others.\\
5) Output ONLY this exact format:
<top>TOP\_CATEGORY</top><sub>SUB\_CATEGORY
</sub>
\end{promptbox}

During the annotation process, images are sampled at a ratio of 100:1 for expert spot-checking, and the final category matching accuracy is 98\%.

After classification is completed, we ask experts to conduct a simple screening of images within each category pair, removing categories with obviously too few samples or generally extremely low image quality. The final dataset contains 92 valid category pairs. The specific coarse-grained and fine-grained categories are shown in Fig.~\ref{fig:pie}. In addition, we especially ask experts to check the images in each other category and remove irrelevant items, in order to prevent Claude from identifying inappropriate frames such as video openings and advertisements as other categories.

To obtain high-quality image samples, we use four reward models: CLIP Score, Aesthetic Score, PickScore, and HPSv3. The texts used for scoring are all image captions annotated by Qwen. The normalization of scores and the calculation process of the total score can be expressed as:
\begin{equation}
\tilde{r}_{m}
=
\frac{
r_{m}-r_{m}^{\min}
}{
r_{m}^{\max}-r_{m}^{\min}
},
\quad m\in\mathcal{M}.
\end{equation}
where $\mathcal{M}=\{\mathrm{clip},\mathrm{aes},\mathrm{hps},\mathrm{pick}\}$.

We rank according to $R_{\mathrm{overall}}$ and take the top 33\% as the filtered real image subset $D_{\mathrm{filter}}$.
\begin{figure*}[t]
    \centering
    \includegraphics[width=1\textwidth]{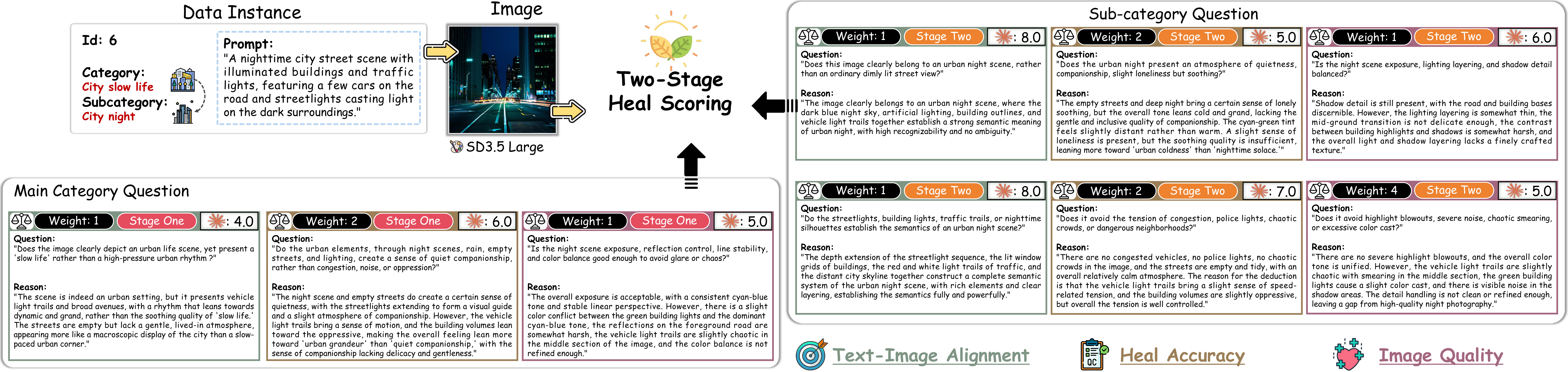}
    \caption{Overview of the two-stage HealScore pipeline. HealScore first evaluates each image with three category-level questions on \textit{text-image alignment}, \textit{heal accuracy}, and \textit{image quality}, and then refines the judgment with six subcategory-level questions. The final score is computed by weighted aggregation of the resulting multi-dimensional responses.}
    \label{fig:healscore}
    \vspace{-15pt}
\end{figure*}

\subsection{Evaluation Criteria and HealScore}
\begin{figure*}[!t]
    \centering
    \includegraphics[width=\textwidth]{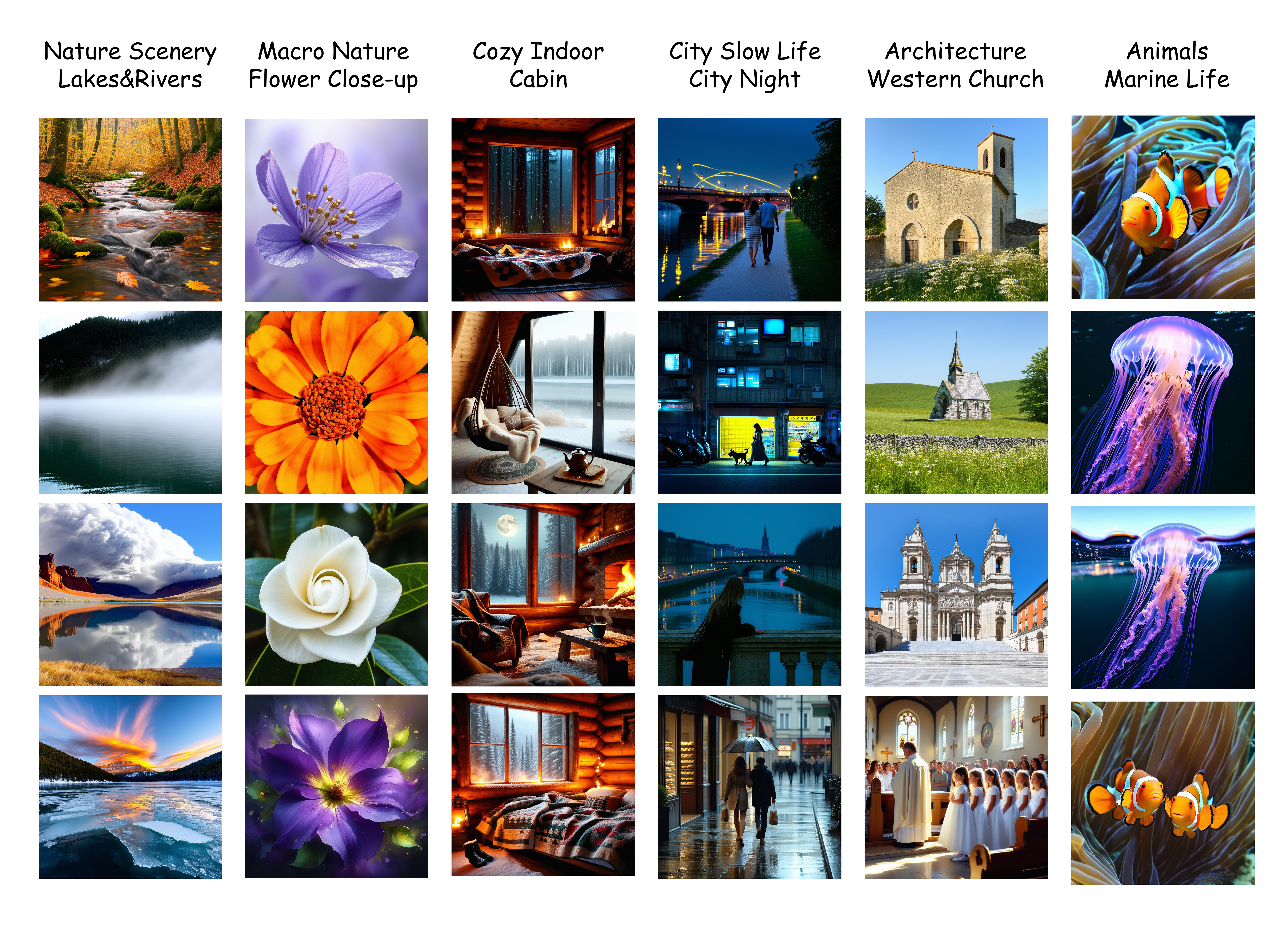}
    \caption{The figure presents augmented images generated by SD3.5-L using variants of the original prompts. The text above each column indicates the top-category and sub-category to which the corresponding images belong.
}
    \label{fig:sup_aug}
\end{figure*}
We model \textbf{HealScore} from three dimensions: \textbf{text-image alignment}, \textbf{heal accuracy}, and \textbf{image quality}, denoted as \texttt{sem}, \texttt{heal}, and \texttt{qual}, respectively. The target question set to be constructed can be defined as:
\begin{equation}
    Q_c = \left\{ q^c_{d,k} \mid d \in \{\texttt{sem}, \texttt{heal}, \texttt{qual}\},\ k \in \{0,1,2\} \right\}.
\end{equation}
Here, $q^c_{d,0}$ denotes the coarse-grained question, while $q^c_{d,1}$ and $q^c_{d,2}$ denote fine-grained supplementary questions.

Each question is first designed by an AI model using the following prompt:
\begin{promptbox}

You are responsible for designing questions for the healing evaluation of image categories.

You need to design questions from three dimensions: text-image alignment, heal accuracy, and image quality.

Text-image alignment refers to whether the image can reflect the healing-oriented semantic features of its corresponding category at the semantic level. Heal accuracy refers to whether the image accurately presents healing-oriented visual features within the given category. Image quality refers to whether the image meets the required quality standard in the healing dimension within the given category.

The top-level category for this design is \{top\_cat\}. Please design one question for this top-level category in each of the three dimensions.

The fine-grained subcategory for this design is \{sub\_cat\}. Please design two questions for this subcategory in each of the three dimensions.

Your final output should strictly follow the JSON format below:
\{\\
\hspace*{1em}"\{top\_cat\}": \{\\
\hspace*{2em}"text\_image\_alignment": "",\\
\hspace*{2em}"heal\_accuracy": "",\\
\hspace*{2em}"image\_quality": ""\\
\hspace*{1em}\},\\
\hspace*{1em}"\{sub\_cat\}": \{\\
\hspace*{2em}"text\_image\_alignment": [\\
\hspace*{3em}"",\\
\hspace*{3em}""\\
\hspace*{2em}],\\
\hspace*{2em}"heal\_accuracy": [\\
\hspace*{3em}"",\\
\hspace*{3em}""\\
\hspace*{2em}],\\
\hspace*{2em}"image\_quality": [\\
\hspace*{3em}"",\\
\hspace*{3em}""\\
\hspace*{2em}]\\
\hspace*{1em}\}\\
\}

\end{promptbox}

After each question is designed, it is reviewed by experts, who either provide revision suggestions as feedback to the AI model or directly modify the question. This process is repeated until all experts agree on the question design.

For each image $I$ in $D_{\mathrm{filter}}$, a large language model scores each question in $q^c_{d,i}$ on a scale from 1 to 10 and provides a detailed rationale. The scoring prompt is as follows:
\begin{promptbox}
Please act as an extremely rigorous ``healing image evaluation expert'' with top-tier aesthetic standards and a professional background in psychology. Please strictly evaluate each of the following questions one by one.

Completely abandon any tolerant attitude of ``close enough is good enough''. Ordinary online images should only receive 4--5 points. Evaluate the image with a highly critical eye, actively look for flaws, and deduct points mercilessly.

Only images that provide exceptionally high aesthetic value or emotional value may receive scores above 8. A score of 10 is reserved only for one-in-a-million perfect artworks.

Please strictly refer to the following score distribution:
- 9--10 points: An absolutely perfect top-tier masterpiece. The image is extremely beautiful, the healing effect touches the soul, and the image quality is flawless.\\
- 7--8 points: Extremely excellent. The subject is extremely clear, the healing atmosphere is strong, and there are no uncomfortable elements.\\
- 4--6 points: Acceptable but ordinary. It meets the basic requirements, but has minor flaws, such as mediocre lighting, slightly rigid composition, or a healing effect that is not deeply moving enough.\\
- 2--3 points (baseline score): A mediocre online image. Although it matches the topic, it lacks aesthetic appeal, has a plain and bland atmosphere, or contains obvious artificial damage, slight noise, or clutter.\\
- 1 point: Poor. The subject is severely misplaced, the image creates a depressing or chaotic feeling, the image quality is poor, or it is severely overexposed/underexposed.\\

Each answer should follow this format: <s>number</s> reason
where the number is an integer from 1 to 10, and the reason should be written in Chinese.
\end{promptbox}

The scored results are further provided to Qwen2.5-VL-7B-Instruct, which serves as the judge model, to assess whether the assigned scores are reasonable, yielding an accuracy of 99\%. The overall scoring pipeline of HealScore is illustrated in Fig.~\ref{fig:healscore}.

After obtaining HealScore following the procedure described in the main text, we select, for each category pair, the top 10 images ranked by HealScore and the top 10 images ranked by the normalized aggregate score of the reward model. For each selected image, Claude rewrites the prompt mainly based on the following prompt:

\begin{promptbox}
You will receive an English description for text-to-image generation/image captioning, hereafter referred to as the original description, together with its coarse-grained category top\_cat and fine-grained category sub\_cat.\\
Task: Under the constraint of strictly adhering to the semantic scope of both the coarse- and fine-grained categories, imitate the tone, sentence length, level of detail, and narrative style of the original description, and write \{n\} new English image descriptions. Each description should resemble a neighboring sample from the same dataset: the subject type should remain consistent, while the specific scene, subject details, actions, or environment should be clearly varied. Do not copy the original sentence or merely replace words with synonyms; do not produce content that is clearly unrelated to the given coarse- or fine-grained category.\\
Language: English; each description should be comparable in length to the original description, although it may be slightly longer or shorter.\\
Output requirements: Do not output any preamble, numbered headings, or Markdown. Output exactly \{n\} English descriptions, separated only by a single line containing --- three hyphens. Apart from this separator, do not include --- anywhere else; do not output more or fewer than \{n\} descriptions.
\end{promptbox}

We denote the resulting augmented prompt as \(C_{\mathrm{augment}}\), which is then used as the input prompt for SD3.5-Large to synthesize additional images, forming the augmented set \(D_{\mathrm{augment}}\). SD3.5-Large uses the pretrained model with 50 inference steps, a guidance-scale coefficient of 7.5, and an output resolution of \(512 \times 512\). To prevent the augmented images from being overly similar to the original images, we compute the similarity between their CLIP embeddings. If the similarity exceeds 0.8, the generated image is discarded, and \(C_{\mathrm{augment}}\) is rewritten to regenerate the image. In Fig.~\ref{fig:sup_aug}, we present the augmented images generated by SD3.5-L based on \(C_{\mathrm{augment}}\).

\subsection{Tri-dimensional Heal Scorer (THS)}

We design and train the Tri-dimensional Healing Scorer (THS) to provide interpretable scoring for image-text samples from three dimensions: semantic relevance, healing property, and image quality. THS uses the BLIP image-text retrieval model as the backbone network to separately encode the input image and text description, and project them into a unified semantic space. Specifically, given an image $I$ and its corresponding text $q$, we first use the visual encoder and text encoder of BLIP to extract image features and text features, respectively, and take the first token from the last-layer sequence representations of both modalities as the global representations. Then, the image representation and text representation are mapped into $256$-dimensional feature vectors through the projection layers of BLIP, denoted as $v$ and $t$. To enhance cross-modal matching capability, we apply $\ell_2$ normalization to $v$ and $t$, and construct the fused feature:
\begin{equation}
h = \mathrm{LayerNorm}([v; t; v \odot t]),
\label{eq:fused_feature}
\end{equation}
where $[\,;\,]$ denotes feature concatenation, and $\odot$ denotes element-wise multiplication. Therefore, the final fused feature $h$ has a dimension of $768$.

The fused feature $h$ is then fed into three independent MLP prediction heads to estimate the scores of semantic relevance, healing property, and image quality. The three prediction heads share the same architecture, each adopting a multilayer perceptron structure of
\begin{equation}
768 \rightarrow 512 \rightarrow 256 \rightarrow 1,
\label{eq:mlp_architecture}
\end{equation}
with GELU activation functions and Dropout regularization in the intermediate layers. The three prediction heads output raw scalars, which are mapped by the Sigmoid function and multiplied by scale coefficients to obtain the scores of the three dimensions:
\begin{equation}
r_{\mathrm{sem}},\quad r_{\mathrm{heal}},\quad r_{\mathrm{qual}}.
\label{eq:dimension_scores}
\end{equation}
Finally, the overall score of THS is obtained by directly summing the scores of the three dimensions:
\begin{equation}
r_{\mathrm{ths}}
=
r_{\mathrm{sem}}
+
2r_{\mathrm{heal}}
+
r_{\mathrm{qual}}.
\label{eq:ths_score}
\end{equation}
Different from directly using the image-text matching classification logits of BLIP, THS only uses BLIP as an image-text feature encoder, and learns fine-grained evaluation criteria for the healing image filtering task through additionally trained tri-dimensional regression heads.

During training, we use the images in $D_{\mathrm{filter}}$ as training samples. For each image, we construct a fixed text template according to its coarse-grained category $T_i$ and fine-grained sub-category $S_{i,j}$:
\begin{equation}
q = \text{``an image of } T_i \ S_{i,j}\text{''}.
\label{eq:text_template}
\end{equation}
Therefore, the inputs of THS are an image $I$ and its category text description $q$, and the outputs are the predicted scores of text-image alignment, heal accuracy, and image quality, as well as the overall score obtained by summing the three scores. Let the supervision targets be
$y_{\text{sem}},y_{\text{heal}}, \text{and} \, y_{\text{qual}},$
with corresponding predictions
$
\hat{y}_{\text{sem}},\hat{y}_{\text{heal}}, \text{and} \, \hat{y}_{\text{qual}}.
$
We first optimize the dimension-wise regression loss:
\begin{equation}
\mathcal{L}_{\text{dim}}
=
\sum_{d\in\{\text{sem},\text{heal},\text{qual}\}}
\left\|\hat{y}_d-y_d\right\|^2.
\end{equation}
To enforce consistency with the summary HealScore while preserving dimension-wise supervision, we further define
\begin{equation}
\begin{aligned}
y_{\text{total}}&=y_{\text{sem}}+2y_{\text{heal}}+y_{\text{qual}},\\
\hat{y}_{\text{total}}&=\hat{y}_{\text{sem}}+2\hat{y}_{\text{heal}}+\hat{y}_{\text{qual}},
\end{aligned}
\end{equation}
and apply an additional total-score loss:
$
\mathcal{L}_{\text{total}}
=
\left\|\hat{y}_{\text{total}}-y_{\text{total}}\right\|^2.
$
The final objective is
$
\mathcal{L}
=
\mathcal{L}_{\text{dim}}
+
\lambda \mathcal{L}_{\text{total}},
$
where \(\lambda\) balances the two terms.

THS is trained in a distributed manner on $6$ GPUs using PyTorch Distributed Data Parallel (DDP). We adopt AdamW as the optimizer, with the weight decay coefficient set to $0.01$. To stabilize the fine-tuning of the pretrained BLIP backbone network, we use different learning rates for the backbone network and the newly added prediction heads. Specifically, the learning rate of the MLP prediction heads and LayerNorm is set to $3\times10^{-4}$, while the learning rate of the BLIP backbone network is set to $0.05$ times the head learning rate, i.e., $1.5\times10^{-5}$. The loss balancing coefficient $\lambda$ is set to $0.2$.

The model is trained for a total of $100$ epochs. The batch size on each GPU is set to $32$, resulting in a global effective batch size of $192$. The learning rate schedule adopts a strategy that combines linear warmup and cosine decay, where the warmup ratio is set to $0.05$. During training, gradient clipping with a maximum gradient norm of $1.0$ is applied, and automatic mixed precision training is enabled to improve training efficiency.

\section{Additional Experimental Settings}
\label{sec:sup-settings}

\begin{figure*}[!t]
    \centering
    \includegraphics[width=\textwidth]{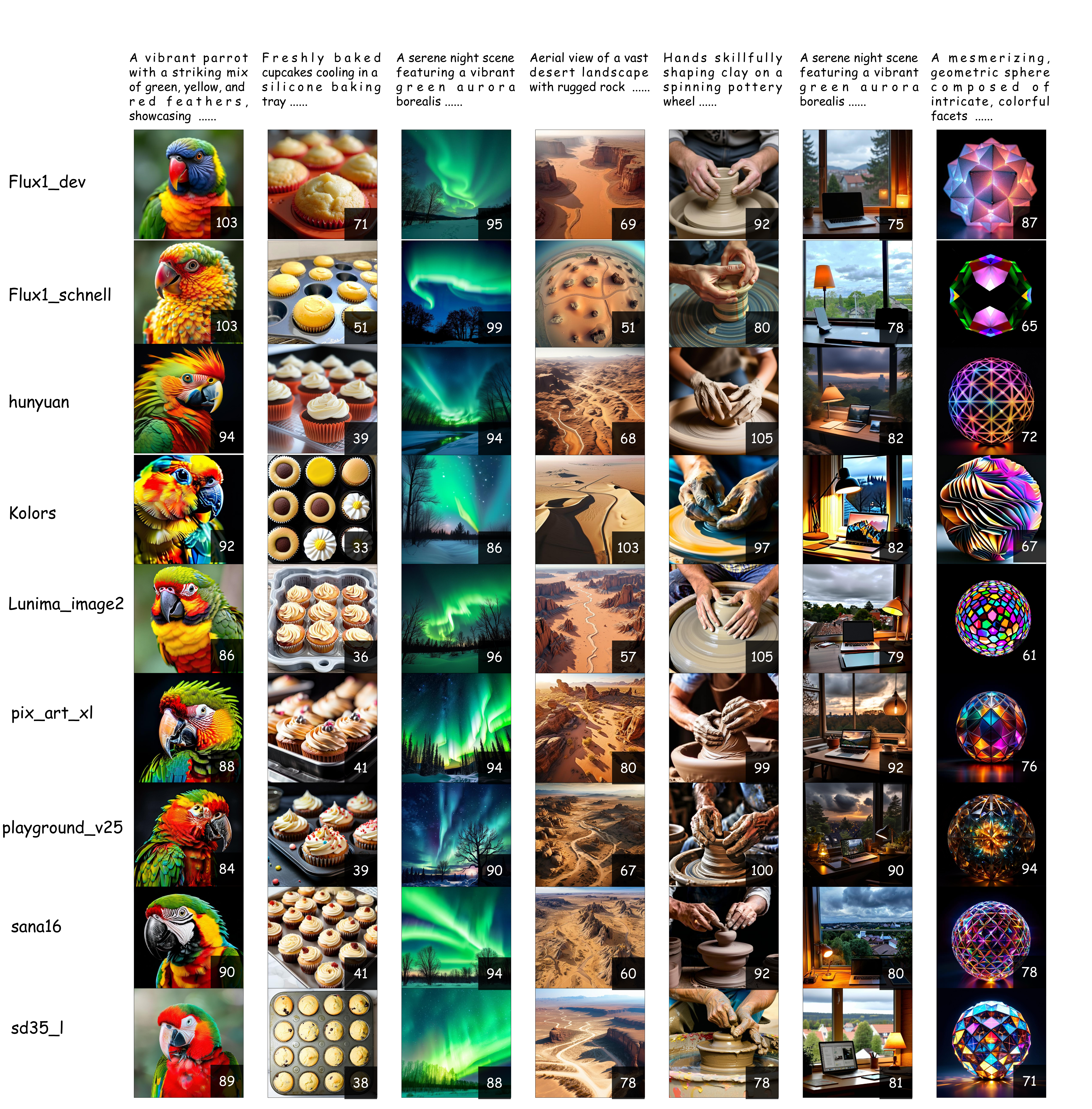}
    \caption{The figure presents representative generation results of several text-to-image baseline models. The number in the bottom-right corner of each generated image denotes the weighted aggregate score of text-image alignment, healing accuracy, and image quality, with a maximum score of 120.
}
    \label{fig:sup_main}
\end{figure*}
\subsection{HealBench-Eval Construction and Validation}
The construction of HealBench-Eval is based on SigLIP and K-Means clustering, and its pseudocode is shown in Algorithm~\ref{alg:test_sampling}.

In the pairwise preference ranking comparison, we first sample $1500$ non-repeated image pairs from HealBench-Eval:
\[
\mathcal{P}
=
\{(I_i^{(a)}, I_i^{(b)})\}_{i=1}^{N},
\quad
N=1500.
\]
Humans and Claude are then asked to separately select the image with stronger healing properties from each pair. Claude-Haiku and Claude-Sonnet 4.6 use the same prompt:

\begin{promptbox}
Please act as a professional evaluator with expertise in healing-oriented and visual-psychological assessment.\\
The following two images are presented in order: the first image corresponds to dataset INDEX \{index\_a\}, and the second image corresponds to INDEX \{index\_b\}.\\
\\
Task: Judge only from the overall perspective of healing quality, perceived psychological comfort, and whether the image can make viewers feel more relaxed and at ease after viewing. Select the image that is better overall.\\
\\
Output requirements, which must be strictly followed:\\
1. Your response must contain one and only one winning-label tag, in the exact format: \textless win\textgreater integer\textless /win\textgreater, where the integer can only be either \{index\_a\} or \{index\_b\}. It must correspond to the image that you consider more healing, where the first image corresponds to \{index\_a\} and the second image corresponds to \{index\_b\}.\\
2. After the tag, write one or two brief sentences explaining the reason. Do not use any other XML tags.\\
3. Do not output any off-topic explanation, numbered list, or Markdown.\\
Example: \textless win\textgreater\{index\_a\}\textless /win\textgreater The color tone is softer, and the scene is more likely to make viewers feel relaxed.
\end{promptbox}

Suppose that for each image pair, the preference result predicted by HealScore is
$y_i^{\mathrm{HS}} \in \{a,b\}$, and the reference preference result obtained from humans or Claude is
$y_i^{\mathrm{ref}} \in \{a,b\}$. Then, the preference agreement accuracy can be expressed as:
\begin{equation}
\mathrm{Acc}
=
\frac{1}{N}
\sum_{i=1}^{N}
\mathbf{1}
\left[
y_i^{\mathrm{HS}} = y_i^{\mathrm{ref}}
\right],
\end{equation}
where $\mathbf{1}[\cdot]$ denotes the indicator function. It takes the value of $1$ when the choice of HealScore is consistent with the choice of humans or Claude, and $0$ otherwise.

Similarly, to further measure the ranking correlation between the score difference of HealScore and the reference preference, we compute Spearman's $\rho$ and Kendall's $\tau$. For the $i$-th image pair, we first define the score difference of HealScore as:
\begin{equation}
\Delta_i^{\mathrm{HS}}
=
s_{i,a}^{\mathrm{HS}}
-
s_{i,b}^{\mathrm{HS}},
\end{equation}
where $s_{i,a}^{\mathrm{HS}}$ and $s_{i,b}^{\mathrm{HS}}$ denote the HealScore of image $a$ and image $b$, respectively. Meanwhile, the reference preference is converted into a signed label:
\begin{equation}
z_i^{\mathrm{ref}}
=
\begin{cases}
+1, & y_i^{\mathrm{ref}} = a,\\
-1, & y_i^{\mathrm{ref}} = b.
\end{cases}
\end{equation}

Spearman's $\rho$ is defined as the Pearson correlation coefficient between the ranked results of the two variables:
\begin{equation}
\rho
=
\frac{
\sum_{i=1}^{N}
\left(
R(\Delta_i^{\mathrm{HS}})-\overline{R_{\Delta}}
\right)
\left(
R(z_i^{\mathrm{ref}})-\overline{R_z}
\right)
}{
\sqrt{
\sum_{i=1}^{N}
\left(
R(\Delta_i^{\mathrm{HS}})-\overline{R_{\Delta}}
\right)^2
}
\sqrt{
\sum_{i=1}^{N}
\left(
R(z_i^{\mathrm{ref}})-\overline{R_z}
\right)^2
}
},
\end{equation}
where $R(\cdot)$ denotes the rank of the corresponding variable, and $\overline{R_{\Delta}}$ and $\overline{R_z}$ denote the mean ranks of
$R(\Delta_i^{\mathrm{HS}})$ and $R(z_i^{\mathrm{ref}})$, respectively.

Kendall's $\tau$ measures ranking consistency by comparing concordant and discordant sample pairs among all sample pairs:
\begin{equation}
\tau
=
\frac{
N_c - N_d
}{
\sqrt{(N_c+N_d+T_{\Delta})(N_c+N_d+T_z)}
},
\end{equation}
where $N_c$ denotes the number of concordant sample pairs, $N_d$ denotes the number of discordant sample pairs, $T_{\Delta}$ denotes the number of sample pairs tied only on
$\Delta^{\mathrm{HS}}$, and $T_z$ denotes the number of sample pairs tied only on $z^{\mathrm{ref}}$.
\begin{algorithm}[t]
\caption{Category-wise Clustering Sampling for HealBench-Eval Construction}
\label{alg:test_sampling}
\begin{algorithmic}[1]
\Require Filtered dataset $D_{\mathrm{filter}}$, captions $C$, threshold $N_{\min}$, number of clusters $K$, samples per cluster $n$
\Ensure Eval dataset $D_{\mathrm{eval}}$

\State $D_{\mathrm{eval}} \gets \emptyset$
\State $\mathcal{G} \gets \emptyset$

\ForAll{$x \in D_{\mathrm{filter}}$}
    \If{$\mathrm{path}(x) \notin C$}
        \State \textbf{continue}
    \EndIf
    \State $q_x \gets C[\mathrm{path}(x)]$
    \State $e_x \gets \mathrm{Normalize}(\mathrm{SigLIP}_{\mathrm{text}}(q_x))$
    \State $\mathcal{G}_{(T_x,S_x)} \gets \mathcal{G}_{(T_x,S_x)} \cup \{(x,e_x)\}$
\EndFor

\ForAll{$\mathcal{G}_{(T,S)} \in \mathcal{G}$}
    \If{$|\mathcal{G}_{(T,S)}| < N_{\min}$}
        \State $D_{\mathrm{eval}} \gets D_{\mathrm{eval}} \cup \{x \mid (x,e_x) \in \mathcal{G}_{(T,S)}\}$
    \Else
        \State $\{\ell_x\}, \{\mu_c\}_{c=1}^{K} \gets \mathrm{KMeans}(\{e_x \mid (x,e_x) \in \mathcal{G}_{(T,S)}\}, K)$
        \For{$c = 1$ to $K$}
            \State $\mathcal{C}_c \gets \{(x,e_x) \in \mathcal{G}_{(T,S)} \mid \ell_x = c\}$
            \If{$|\mathcal{C}_c| \le n$}
                \State $D_{\mathrm{eval}} \gets D_{\mathrm{eval}} \cup \{x \mid (x,e_x) \in \mathcal{C}_c\}$
            \Else
                \State $\mathcal{S}_c \gets$ the $n$ samples in $\mathcal{C}_c$ with the smallest $\|e_x-\mu_c\|_2^2$
                \State $D_{\mathrm{eval}} \gets D_{\mathrm{eval}} \cup \mathcal{S}_c$
            \EndIf
        \EndFor
    \EndIf
\EndFor

\State \Return $D_{\mathrm{eval}}$
\end{algorithmic}
\end{algorithm}

\subsection{Benchmarking Baseline T2I Models}
For the T2I baseline models, we uniformly use the image descriptions from HealBench-Eval as prompt inputs. The basic configuration of each model is designed according to its official documentation, and the specific parameters are shown in Table~\ref{tab:generation_settings}.

\begin{table}[t]
\centering
\caption{Generation settings of evaluated T2I baseline models.}
\label{tab:generation_settings}
\begin{tabular}{lccc}
\toprule
Model & Steps & Guidance Scale & Resolution \\
\midrule
SD 2.1 & 50 & 7.5 & $512 \times 512$ \\
SDXL Base 1.0 & 50 & 7.5 & $512 \times 512$ \\
PixArt-XL & 50 & 7.5 & $512 \times 512$ \\
SDXL Turbo & 50 & 1.0 & $512 \times 512$ \\
Playground v2.5 & 50 & 7.5 & $1024 \times 1024$ \\
Hunyuan-DiT & 50 & 7.5 & $512 \times 512$ \\
Kolors & 50 & 7.5 & $512 \times 512$ \\
Sana-1.6B & 50 & 4.5 & $512 \times 512$ \\
SD 3 Medium & 50 & 7.5 & $512 \times 512$ \\
FLUX.1 [dev] & 50 & 3.5 & $512 \times 512$ \\
FLUX.1 [schnell] & 50 & 0 & $512 \times 512$ \\
SD 3.5 Large & 50 & 7.5 & $512 \times 512$ \\
SD 3.5 Medium & 50 & 7.5 & $512 \times 512$ \\
Lumina-Image 2.0 & 50 & 7.5 & $512 \times 512$ \\
\bottomrule
\end{tabular}
\end{table}

After image generation, we input the generated images into Claude and ask it to score each image according to the nine questions corresponding to its category pair. To prevent different image resolutions from affecting the evaluation, we proportionally resize all images to a resolution of $512 \times 512$ before input. The generated images and corresponding scores of these baseline T2I models evaluated on HealBench are shown in Fig.~\ref{fig:sup_main}.

\subsection{Prompt Rewriting as Intervention}
\begin{figure*}[!t]
    \centering
    \includegraphics[width=\textwidth]{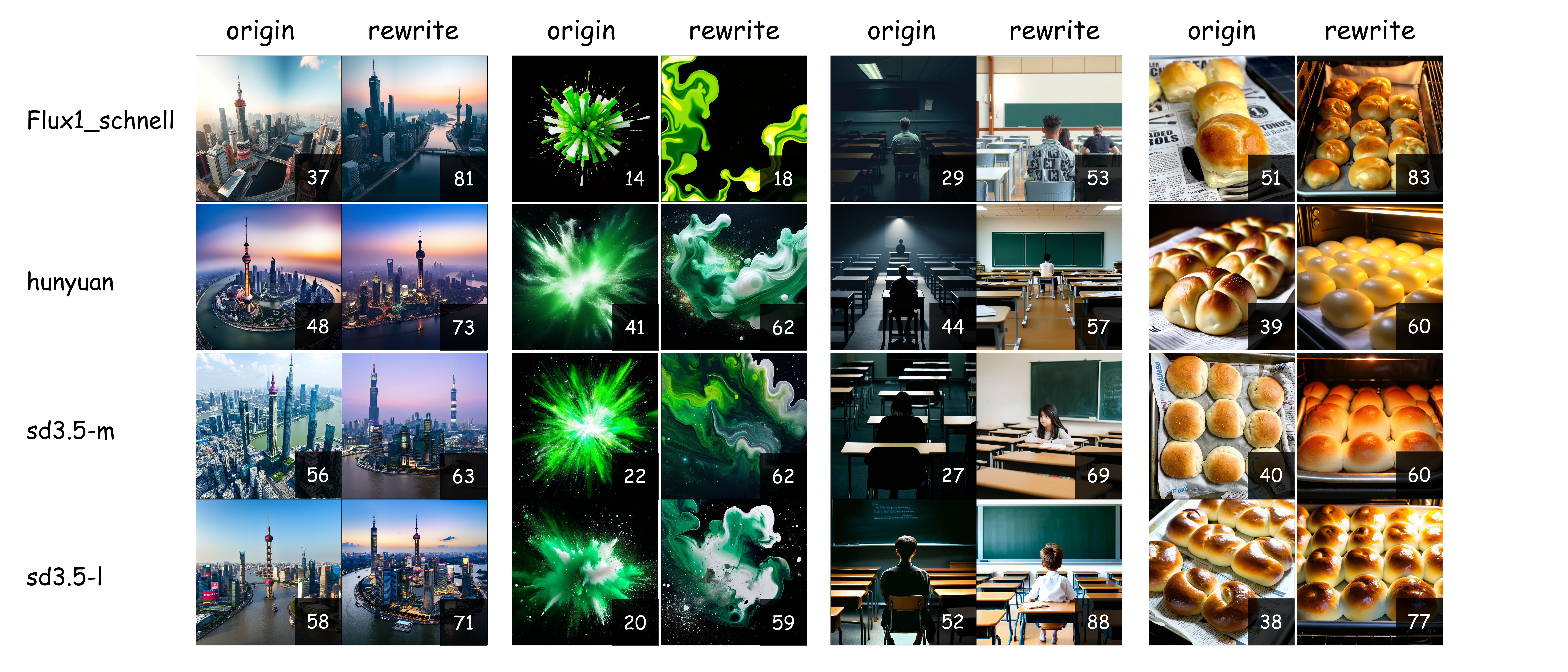}
    \caption{The figure illustrates the effectiveness of our prompt rewriting strategy. We evaluate the strategy using four text-to-image models. In each image pair, the left image is generated from the original prompt, while the right image is generated from the prompt rewritten based on Claude's evaluation. The number in the bottom-right corner denotes the weighted aggregate score.
}
    \label{fig:sup_re}
\end{figure*}

\begin{figure*}[!t]
    \centering
    \includegraphics[width=\textwidth]{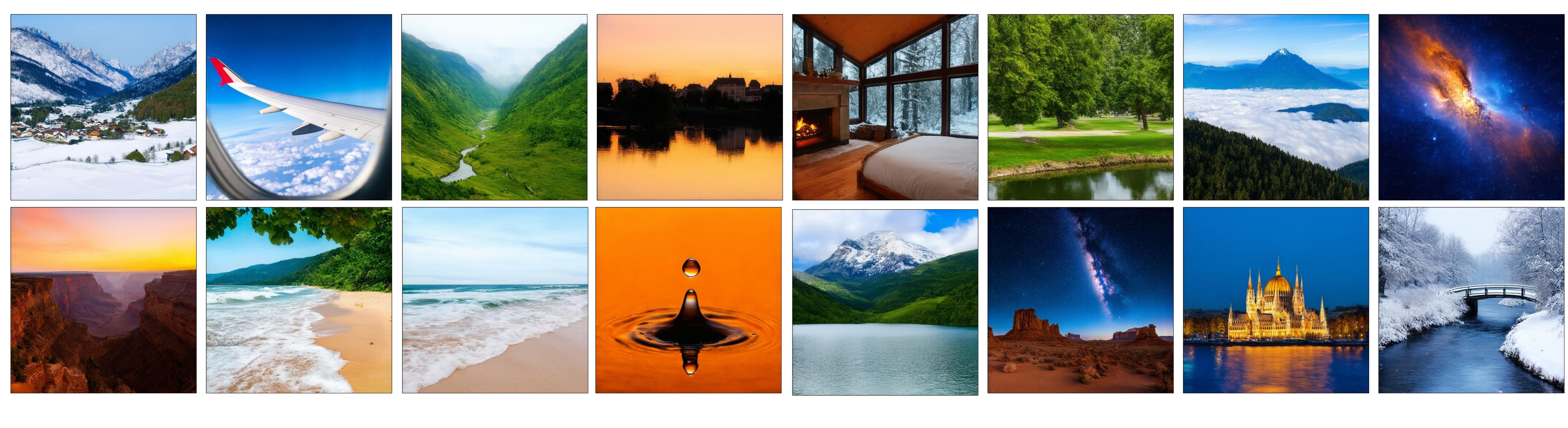}
    \caption{The figure presents images generated from the original prompts in HealBench using SD3.5-M fine-tuned with FLOW-GRPO.
}
    \label{fig:sup_RL}
\end{figure*}
During each scoring process in which Claude is asked to evaluate an image, we require it to provide both a score and a rationale for each corresponding question. In the improvement strategy based on Prompt Rewriting, we feed Claude the scores and scoring rationales of a given image across different dimensions, and ask it to rewrite the caption according to the following basic rules:

\begin{promptbox}
You will receive: the image path, the coarse-grained category top\_cat, the fine-grained category sub\_cat, the original English Caption, and the experts' integer scores and comments (reason) for this image across multiple dimensions.\\
Task: Based on the above information, rewrite the Caption so that its expression becomes reasonably more consistent with the expert comments, while the core subject, scene, and focus described in the original Caption must be preserved. The rewritten description should therefore exhibit higher heal accuracy and perceived psychological comfort.\\
Do not change the topic, do not introduce content that contradicts either the comments or the original Caption, and do not fabricate objects or events that are absent from the image.\\
The rewriting language should be consistent with the original Caption, which is usually English.\\
Output requirement: output only one line, and strictly follow the format below without any additional explanation or label:\\
\textless caption\textgreater The complete rewritten Caption text \textless /caption\textgreater
\end{promptbox}

We denote the resulting rewritten image description as $C_{\mathrm{rewrite}}$. We select four representative text-to-image (T2I) models and use the rewritten $C_{\mathrm{rewrite}}$ as input to examine and score the generation results after rewriting. The basic generation settings of these models still follow those reported in Table~\ref{tab:generation_settings}.

In addition, we also use the description rewriting strategy to explore the potential improvement of Image Edit models. Specifically, we use the original image before editing as the base image input, and use $C_{\mathrm{rewrite}}$ as the image editing prompt. Note that we forcibly normalize the resolution of the original image to $512 \times 512$ through cropping or resizing. We adopt Pix2Pix and SDXL-Pix2Pix as the image editing models. For both models, the output resolution is set to $512 \times 512$, and the number of denoising steps is set to 50. In Fig.~\ref{fig:sup_re}, we present several generated images obtained through the Prompt Rewriting strategy, along with a comparison of their scores against the corresponding original images. It can be observed that the rewritten prompts effectively help T2I models generate images with enhanced healing attributes.

\subsection{THS as an Automatic Healing Scorer}
In the evaluation of THS, we also adopt pairwise preference ranking. For the sampled $1500$ non-repeated preference pairs, denoted as
\begin{equation}
\mathcal{Q}
=
\{(I_i^{(a)}, I_i^{(b)})\}_{i=1}^{N_{\mathrm{ths}}},
\quad
N_{\mathrm{ths}}=1500,
\end{equation}
let $y_i^{\mathrm{CS}}\in\{a,b\}$ denote the reference preference obtained from the summed Claude-Sonnet scores, and let $y_i^{\mathrm{M}}\in\{a,b\}$ denote the preference obtained from THS or other general-purpose scoring models. The preference accuracy of a scoring model is defined as:
\begin{equation}
\mathrm{Acc}_{\mathrm{M}}
=
\frac{1}{N_{\mathrm{ths}}}
\sum_{i=1}^{N_{\mathrm{ths}}}
\mathbf{1}
\left[
y_i^{\mathrm{M}} = y_i^{\mathrm{CS}}
\right].
\end{equation}

To compute ranking correlations, we further define the score difference of the evaluated model as:
\begin{equation}
\Delta_i^{\mathrm{M}}
=
s_{i,a}^{\mathrm{M}}
-
s_{i,b}^{\mathrm{M}},
\end{equation}
and convert the Claude-Sonnet reference preference into a signed label:
\begin{equation}
z_i^{\mathrm{CS}}
=
\begin{cases}
+1, & y_i^{\mathrm{CS}} = a,\\
-1, & y_i^{\mathrm{CS}} = b.
\end{cases}
\end{equation}
Then, following the definitions introduced above, we compute Kendall's $\tau$ and Spearman's $\rho$ between
$\{\Delta_i^{\mathrm{M}}\}_{i=1}^{N_{\mathrm{ths}}}$
and
$\{z_i^{\mathrm{CS}}\}_{i=1}^{N_{\mathrm{ths}}}$.
\subsection{Exploratory RL Fine-Tuning with THS Reward}


In Flow-GRPO training, we use SD3.5-medium as the base text-to-image model and perform parameter-efficient fine-tuning with LoRA, where the rank is set to $32$. EMA is also enabled to stabilize the training process. All images are generated and evaluated at a unified resolution of $512 \times 512$. The number of sampling steps is set to $10$ during training and $40$ during evaluation, with the CFG guidance scale set to $4.5$. For each prompt, $4$ images are generated. The sampling batch size per GPU is set to $2$, and each epoch contains $2$ sampling batches. During training, the batch size per GPU is set to $2$, and the gradient accumulation step is set to $1$. 

During optimization, we adopt a PPO/GRPO-style KL constraint, with the KL coefficient $\beta$ set to $0.04$, and use fp16 mixed-precision training to improve training efficiency. The reward function is set to THS, with reward weights of $1:2:1$ for text-image alignment, heal accuracy, and image quality, respectively. In addition, we enable per-prompt statistic tracking to normalize the reward distribution under different prompts, thereby improving training stability. In Fig.~\ref{fig:sup_RL}, we present several images generated by the SD3.5-m model after reinforcement learning.
\section{Human Evaluation Protocol and Ethical Considerations}
\label{sec:sup-human}

This section provides additional details on the design, participant recruitment, experimental procedure, outcome assessment, participant protection, and ethical governance of the preliminary human study reported in the main paper. The study was designed to examine whether the six HEAL classes could support preference-sensitive delivery of psychologically comforting visual content. Specifically, we compared a personalized condition, in which participants selected a preferred HEAL class, with a non-personalized condition, in which healing-oriented images were presented without participant-directed class selection. The study was exploratory and was not designed to establish clinical efficacy or to replace professional psychological assessment, diagnosis, or treatment.

\subsection{Study Design}
\label{sec:sup-human-design}

We conducted a preliminary between-condition human study involving 24 college students with subthreshold depressive symptoms. Participants were allocated to two conditions, with 12 participants in the random-playback condition and 12 participants in the personalized class-selection condition. The allocation and study procedures followed the protocol approved by the responsible institutional ethics committee.

Participants in the random-playback condition viewed healing-oriented images selected without reference to their individual HEAL-class preferences. Participants in the personalized class-selection condition were allowed to choose among the six HEAL classes: Natural Restoration, Atmospheric Immersion, Shelter and Safety, Living Companionship, Rhythmic Focus, and Awe and Transcendence. Images were then presented from the selected class. Except for the class-selection component, the two conditions followed the same general presentation and assessment procedure.

The study examined short-term responses during a controlled image-viewing session. It did not involve psychotherapy, medication, clinical diagnosis, or modification of participants' existing medical or psychological care. The visual materials were treated as experimental stimuli rather than as therapeutic products.

\subsection{Participants and Eligibility}
\label{sec:sup-human-participants}

Participants were recruited from a university population according to the eligibility criteria specified in the approved study protocol. Eligible participants were college students who met the predefined operational criteria for subthreshold depressive symptoms, were able to understand the study instructions, and voluntarily agreed to participate.

Eligibility was assessed using the screening and assessment procedures defined in the approved protocol. Participants whose condition exceeded the intended scope of a low-risk image-viewing study, including individuals requiring immediate clinical assessment or crisis intervention, were not enrolled in the experimental session. Additional identifying information concerning the recruitment site and responsible institution is omitted from the anonymized manuscript to preserve double-blind review.

Participation was voluntary. Recruitment materials and study personnel did not characterize the experiment as a substitute for psychological treatment, nor did they promise clinical improvement as a consequence of participation.

\subsection{Visual Stimuli}
\label{sec:sup-human-stimuli}

The visual stimuli were selected from the healing-oriented image collection developed in this work and were organized according to the six HEAL classes. The stimuli were intended to represent different restorative functions, including nature-based recovery, immersive atmosphere, perceived shelter and safety, affiliative companionship, calm process-oriented attention, and experiences of awe or transcendence.

Before their use in the human study, the images underwent the data-quality and healing-oriented filtering procedures described in the main paper. Images containing explicit violence, self-harm content, graphic medical content, clearly threatening scenes, or other overtly distressing material were excluded. Because emotional responses to visual content remain individually variable, the study did not assume that every image would be experienced as comforting by every participant.

The two experimental conditions used the same general source of visual materials. Their principal difference was whether the participant could select the HEAL class from which the images were presented.

\subsection{Experimental Procedure}
\label{sec:sup-human-procedure}

Before the experiment, participants received a description of the study purpose, procedures, expected duration, types of data collected, foreseeable discomforts, privacy protections, and their right to withdraw. Participants provided informed consent before any experimental procedure or study-specific assessment was conducted.

Participants first completed the baseline assessment specified in the approved protocol. They then entered their assigned image-viewing condition. In the personalized class-selection condition, participants reviewed descriptions of the six HEAL classes and selected the class that best matched their current preference or desired visual experience. Images belonging to the selected class were subsequently presented. In the random-playback condition, images were presented without allowing participants to choose a HEAL class.

The viewing session lasted approximately one hour. The presentation environment and general instructions were kept consistent across conditions to the extent permitted by the study design. Participants were instructed to view the images naturally and were informed that they could pause or discontinue the session if they experienced discomfort.

After the image-viewing session, participants completed the post-session assessment. The resulting pre-session and post-session measurements were used to describe short-term changes associated with the two content-delivery conditions.

\subsection{Outcome Assessment}
\label{sec:sup-human-outcomes}

The study reported two primary descriptive outcomes: the proportion of participants classified as showing improvement and the change in HAMD-17 score between the pre-session and post-session assessments. The operational definition of improvement, the assessment procedure, and the handling of incomplete observations followed the approved study protocol.

HAMD-17 assessments were administered according to the study's predefined assessment procedure by personnel who had received the training required under the approved protocol. The same assessment framework was applied across the two experimental conditions.

As reported in Table~\ref{tab:user_study_delivery}, the personalized class-selection condition exhibited a higher improvement ratio and a larger average reduction in HAMD-17 score than the random-playback condition. Given the small sample size and short observation period, these outcomes were interpreted descriptively as preliminary feasibility evidence. They were not interpreted as demonstrating sustained symptom reduction, treatment efficacy, or a clinically established therapeutic effect.

\subsection{Analysis and Reporting}
\label{sec:sup-human-analysis}

The human-study results were summarized at the condition level using the number of participants, the proportion classified as showing improvement, and the average pre-session to post-session change in HAMD-17 score. The study was not powered as a confirmatory clinical trial, and the reported analysis was intended to characterize the direction and magnitude of the observed short-term differences rather than to support definitive clinical inference.

The personalized and random conditions were compared under the same reporting framework. No result was interpreted independently of the exploratory nature of the study, the restricted participant population, or the short duration of the intervention.

Any missing or incomplete observations were handled according to the approved analysis protocol. The denominators used for the reported proportions correspond to the participants with outcome information included in the relevant analysis.

\subsection{Informed Consent and Voluntary Participation}
\label{sec:sup-human-consent}

All participants provided informed consent before participation. The consent process explained the purpose of the research, the experimental procedures, the expected duration, the types of data collected, foreseeable risks and discomforts, potential benefits, privacy protections, and the intended use of the research findings.

Participants were explicitly informed that participation was voluntary and that they could decline to answer a question, pause the viewing session, or withdraw from the study at any time without penalty. Withdrawal did not affect participants' academic status, access to institutional services, or access to psychological support.

The consent materials did not describe the visual stimuli as a clinically validated treatment and did not guarantee that participation would improve mood or depressive symptoms.

\subsection{Participant Safety and Risk Management}
\label{sec:sup-human-safety}

The study was designed as a low-risk image-viewing experiment. The visual stimuli were curated to emphasize restorative and psychologically comforting content and to exclude explicit violence, self-harm imagery, graphic injury, and other clearly distressing material.

Nevertheless, because individual emotional responses may differ, participants were instructed that they could stop viewing any image, pause the experiment, or discontinue participation if they experienced discomfort. Study personnel followed the safety-monitoring and escalation procedures defined in the approved protocol.

Participants whose screening or study participation indicated a level of psychological risk beyond the scope of the experiment were managed according to the approved risk-assessment and referral procedure. Where appropriate, participants were directed to qualified psychological or medical support services. The experimental procedure was not used to make an independent clinical diagnosis or emergency-care decision.

\subsection{Privacy and Data Protection}
\label{sec:sup-human-privacy}

Only data necessary for the research objectives were collected. Each participant was assigned a research identifier, and directly identifying information was stored separately from experimental responses and assessment records. The analyses reported in the paper were conducted using de-identified data.

Access to participant-level data was restricted to authorized members of the research team. Data storage, retention, access control, and deletion followed the approved protocol and the applicable institutional requirements. No directly identifying information is included in the manuscript, supplementary material, tables, or qualitative descriptions.

Participant-level data will not be publicly released unless such release is explicitly permitted by the informed-consent agreement, the ethics approval, and applicable privacy requirements. Any future data sharing will use an appropriately de-identified or aggregated form.

\subsection{Compensation}
\label{sec:sup-human-compensation}

Participants received a fixed amount approved by the ethics committee. Compensation was not contingent on providing a particular response, demonstrating improvement, completing optional assessments, or remaining in the study after requesting withdrawal. The level and form of compensation were designed not to constitute undue inducement.

\subsection{Ethics Approval}
\label{sec:sup-human-ethics}

The human-participant component of this study was reviewed and approved by the responsible institutional ethics committee before participant recruitment and data collection. The study was conducted in accordance with the approved protocol, applicable institutional requirements, and established ethical principles for research involving human participants.

To preserve anonymity during double-blind review, the name of the approving institution, the name of the ethics committee, the ethics approval number, and other institution-identifying information are withheld from the anonymized manuscript. These details will be restored in the de-anonymized version of the paper.

The omission of these identifiers from the anonymized submission reflects the double-blind review policy and does not indicate that ethical approval was pending or unavailable.

\subsection{Trial Registration}
\label{sec:sup-human-registration}

The broader study record is registered with the \underline{XXXX} Clinical Trial Registry. The registration number is masked in the anonymized submission as \underline{\textbf{CXXXTR2600117XXX}} (specific details will be disclosed after this paper is accepted). The complete registration identifier and corresponding study information will be disclosed after de-anonymization in accordance with the conference's review policy.

Trial registration and institutional ethics approval represent separate components of research governance. The registry provides a formal record of the broader study, whereas authorization to conduct the human-participant procedures was granted by the responsible institutional ethics committee.

\subsection{Scope and Limitations of the Human Study}
\label{sec:sup-human-limitations}

The human-study findings should be interpreted within several boundaries. First, the study involved a small and specific sample of college students with subthreshold depressive symptoms. Second, it examined short-term responses to a single image-viewing session rather than long-term psychological outcomes. Third, the personalized condition supported selection at the HEAL-class level and did not model the full range of individual clinical, cultural, situational, or visual preferences that may influence perceived psychological comfort.

Accordingly, the study provides preliminary evidence that the six-class HEAL taxonomy may support preference-sensitive delivery of restorative visual content. It does not establish that HealBench images constitute a clinical intervention, that image exposure treats depression, or that the observed differences generalize to clinical populations or long-term use.

Larger samples, confirmatory and preregistered designs, independently replicated protocols, longer follow-up periods, and clinically appropriate outcome assessments would be required before making therapeutic or clinical claims.
